\documentclass{article}

\usepackage{microtype}
\usepackage{graphicx}
\usepackage{subfigure}
\usepackage{booktabs} 

\usepackage{hyperref}

\usepackage{amsmath}
\usepackage{listings}
\usepackage{color}
\usepackage{times}
\usepackage{soul}
\usepackage{url}
\usepackage[utf8]{inputenc}
\usepackage[small]{caption}
\usepackage{algorithmic}
\usepackage{xcolor}
\usepackage{etoc}
\usepackage{wrapfig}
\usepackage{tabularx}
\usepackage{adjustbox}
\usepackage{bbm}
\usepackage{multicol}
\usepackage{multirow}

\definecolor{dkgreen}{rgb}{0,0.6,0}
\definecolor{gray}{rgb}{0.5,0.5,0.5}
\definecolor{mauve}{rgb}{0.58,0,0.82}

\definecolor{darkgoldenrod}{rgb}{0.72, 0.53, 0.04}
\definecolor{indianred}{rgb}{0.8, 0.36, 0.36}
\definecolor{mediumseagreen}{rgb}{0.24, 0.7, 0.44}
\definecolor{mediumpurple}{rgb}{0.58, 0.44, 0.86}

\usepackage[accepted]{icml2025}

\usepackage{amsmath}
\usepackage{amssymb}
\usepackage{mathtools}
\usepackage{amsthm}

\usepackage[capitalize,noabbrev]{cleveref}

\theoremstyle{plain}

\theoremstyle{definition}

\theoremstyle{remark}

\usepackage[textsize=tiny]{todonotes}

\icmltitlerunning{Code Simulation as a Proxy for High-order Tasks in Large Language Models}

\begin{document}

\twocolumn[
\icmltitle{Code Simulation as a Proxy for High-order Tasks in Large Language Models}



\icmlsetsymbol{equal}{*}

\begin{icmlauthorlist}
\icmlauthor{$^*$Emanuele La Malfa}{ox,tur}
\icmlauthor{Christoph Weinhuber}{ox}
\icmlauthor{Orazio Torre}{sal}
\icmlauthor{Fangru Lin}{oxlin}
\icmlauthor{X. Angelo Huang}{eth}
\icmlauthor{Samuele Marro}{oxeng}
\icmlauthor{Anthony Cohn}{tur,leeds}
\icmlauthor{Nigel Shadbolt}{ox,tur}
\icmlauthor{Michael Wooldridge}{ox}
\end{icmlauthorlist}

\icmlaffiliation{ox}{Department of Computer Science, University of Oxford}
\icmlaffiliation{oxlin}{Faculty of Linguistics, Philology, and Phonetics, University of Oxford}
\icmlaffiliation{oxeng}{Department of Engineering, University of Oxford}
\icmlaffiliation{eth}{Department of Computer Science, ETH Zurich}
\icmlaffiliation{leeds}{Faculty of Engineering and Physical Sciences, University of Leeds}
\icmlaffiliation{sal}{University of Salerno}
\icmlaffiliation{tur}{The Alan Turing Institute}

\icmlcorrespondingauthor{Emanuele La Malfa}{emanuele.lamalfa@cs.ox.ac.uk}

\vskip 0.2in
\centering
{\includegraphics[height=1.2em]{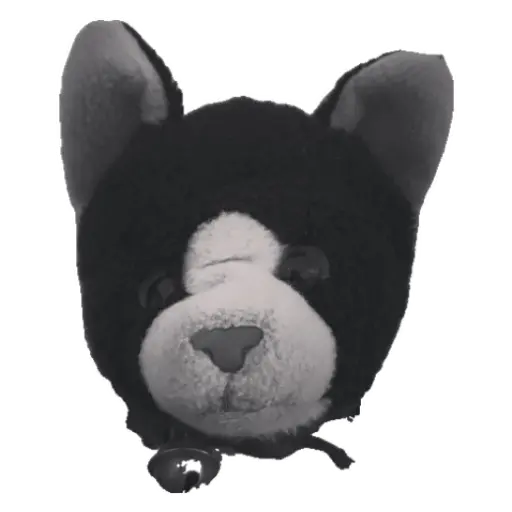} \href{https://emanuelelm.github.io/projects/codesim/}{Project Website}}
\vskip 0.1in

\icmlkeywords{Machine Learning, ICML}

\vskip 0.3in
]



\printAffiliationsAndNotice{}  

\begin{abstract}
Many reasoning, planning, and problem-solving tasks share an intrinsic algorithmic nature: correctly simulating each step is a sufficient condition to solve them correctly.
We collect pairs of naturalistic and synthetic reasoning tasks to assess the capabilities of Large Language Models (LLM).
While naturalistic tasks often require careful human handcrafting, we show that synthetic data is, in many cases, a good proxy that is much easier to collect at scale.
We leverage common constructs in programming as the counterpart of the building blocks of naturalistic reasoning tasks, such as straight-line programs, code that contains critical paths, and approximate and redundant instructions. We further assess the capabilities of LLMs on sorting problems and repeated operations via sorting algorithms and nested loops.
Our synthetic datasets further reveal that while the most powerful LLMs exhibit relatively strong execution capabilities, the process is fragile: it is negatively affected by memorisation and seems to rely heavily on pattern recognition.
Our contribution builds upon synthetically testing the reasoning capabilities of LLMs as a scalable complement to handcrafted human-annotated problems.
\end{abstract}

\section{Introduction}
\begin{figure*}
\centering
\includegraphics[width=0.85\linewidth]{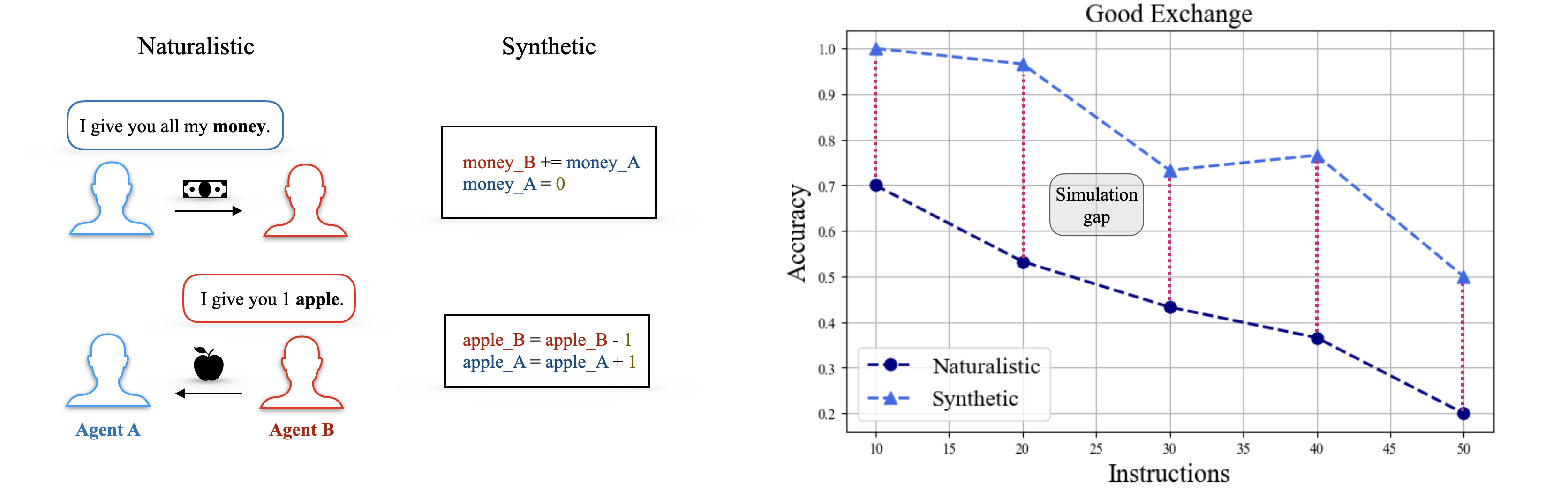}
\caption{Left: an example of the naturalistic vs. synthetic good exchange settings. The former describes, in natural language, two agents who exchange goods; the latter is an equivalent formulation in code. While GPT-3.5-Turbo performs better on the synthetic task (a ``simulation gap''), performance in the synthetic and naturalistic tasks is strongly correlated with respect to the control variable, i.e., the number of operations/exchanges. We conduct experiments on $30$ samples per instruction class with $\{10, 20, 30, 40, 50\}$ interactions/ lines of code. 
}
\label{fig:motivating-example}
\vspace{-3mm}
\end{figure*}

A major area of interest at the time of writing is understanding the capabilities of Large Language Models (LLMs) beyond the tasks of language understanding and generation. 
Many benchmarks, including Theory of Mind~\cite{pmlr-v80-rabinowitz18a}, planning~\cite{hao2023reasoning,ouyang2022training}, and high-order reasoning~\cite{webb2023emergent}, necessitate turning a prompt, expressed in natural language, into a procedure that must then be carried out faultlessly.  
Consider a problem where two agents interact and exchange goods, as in Figure~\ref{fig:motivating-example}. 
An LLM prompted to compute the number of goods at the end of an iteration should be able to sum and assign variables correctly. Such a naturalistic problem has an intrinsic algorithmic nature as code (Figure~\ref{fig:motivating-example}, centre). This work centres on this analogy and turns it into an experimental pipeline to assess the capabilities of LLMs on reasoning tasks via equivalent code simulation.

Our preliminary experiments, which we will expand on in the next sections and pair with others in literature~\cite{lin2024graphenhanced}, evidence a strong performance correlation between naturalistic and synthetic tasks. 
Beyond the example above, the code simulation capabilities of LLMs are an important object of study for at least two complementary reasons. First, LLMs have shown planning capabilities~\cite{ouyang2022training}, which requires reasoning step-by-step and recursively dividing a problem into its elementary components. However, such compositional reasoning performance is highly fragile to variations~\cite{turpin2024language}. With highly structured input, code simulation can shed insight into the underlying failure modes of the LLM without the confounding factors of natural language tasks.
Second, code simulation requires an LLM to turn instructions formulated in code and natural language into a procedure the model must solve correctly, and thus answers the question of whether LLMs can serve as \emph{digital} computational models~\cite{jojic2023gpt}.

This paper shows that code simulation is a scalable proxy for assessing some core reasoning capabilities of LLMs; this is in contrast with generating naturalistic datasets, which is resource- and time-expensive.
While not all the reasoning tasks can be equivalently formulated as code, we showcase the flexibility of our framework by pairing five non-trivial naturalistic tasks with their coding counterpart and show that the performances of GPT-4, GPT-4o~\cite{OpenAI2023GPT4TR}, and Llama3.1-405B~\cite{grattafiori2024llama3herdmodels} correlate with them.
As expected, our code simulation approach enables us to identify failure cases for step-by-step execution, namely memorisation and ``lazy'' pattern recognition. 
For these cases, we develop a minimal extension of Chain of Thought~\cite{wei2022chain} that mitigates these issues and allows benchmarking in the presence of memorisation.
To further lay the groundwork for code simulation as a proxy of naturalistic reasoning tasks, we conduct extensive experiments on pure code simulation and simple programming constructs on a variety of models, including GPT-3.5-Turbo, GPT-4, Jurassic2-Ultra, Llama-2-70B, Llama-3-70B and CodeLLaMA-34b-Instruct. The code to replicate the experiments in the main paper is available \href{https://github.com/EmanueleLM/naturalistic-datasets-codesim}{here}, while the code to replicate the experiments on pure code simulation is available \href{https://github.com/EmanueleLM/CodeSimulation}{here}.
Click \href{https://emanuelelm.github.io/projects/codesim/}{here} to access the project's website.

\section{Related Work} 
LLMs to understand, generate, and improve code have been mainly developed to produce debugging information without invoking a compiler/interpreter~\cite{hou2023large,santos2023always,chen2021evaluating,widjojo2023addressing,zan2023large}.
Code generation and simulation require some degree of compositionality~\cite{mccoy2023much}, i.e., the result of complex expressions can be determined by their constituents and the rules used to combine them.
Recent works explored compositionality in terms of simple mathematical operations that LLMs can execute~\cite{frieder2023mathematical,yang2023code,yuan2023well}, and revealed how the most potent models do not achieve that~\cite{mccoy2023embers,west2023generative}.
Our work further explores the tension between memorisation and performance on complex tasks~\cite{berglund2023reversal,eldan2023s,yang2023code}, with results that illustrate how the former is at tension with the size of a model, the so-called ``inverse scaling law''~\cite{biderman2023emergent}.

Before the breakthrough of closed-source LLMs~\cite{la2023arrt}, a seminal work tested LLMs on code simulation, showing how keeping track of the variables improves their capabilities~\cite{nye2021show}. Successive works explored LLMs and code simulation~\cite{chen2024language,tufano2023predicting,zhou2023algorithms}, particularly in~\cite{liu2023code}, where the authors fine-tune Transformer-based models to output the program trace of a code snippet. 
The code simulation capabilities of LLMs have been explored in several recent works: the first that identified the problem as relevant for LLMs is~\cite{la2024code}, of which this work is an extension. Other works successively extended this idea~\cite{lyu2024largelanguagemodelscode}.
Recent developments in this field go under the name of ``code reasoning'', as a model's ability to predict a variable's state at runtime~\cite{chen2024evaluating}, the output of a statement/function~\cite{gu2024cruxeval,liu2024codemind}, or their capability to handle recursion~\cite{zhang2024transformerbased}.
At the architectural level, several works studied Transformers and attention-based models regarding the operations and programming languages they interpret and execute~\cite{weiss2021thinking} and their recursive code simulation capabilities~\cite{zhang2023can}.
On a broader perspective, past work investigated the Turing-completeness of LLMs~\cite{giannou2023looped,perez2021attention,schuurmans2023memory,wei2022statistically}, and their ability to follow instructions~\cite{ouyang2022training} and policies expressed as code~\cite{liang2023code}.

\section{Methodology}\label{sec:methodology}
Formally, a Language Model is defined as a function that predicts the next token (out of a finite vocabulary) conditioned on the sequence of previously fed/generated tokens, namely $\psi: V^* \xrightarrow{} \mathbb{P}(V)$. 
In our setting, a problem is specified as a tuple $(x, p)$, where $x$ instructs the model to solve a problem $p$ expressed either as code or as an equivalent naturalistic task.
Both settings express the same question, with the ground truth label obtained by running the code with an interpreter/compiler $\Omega$ and compared, for correctness, against the model's answer.
We select Python 3 as our programming language for coding problems as it represents the language of reference in most LLMs, such as Code-LLaMA~\cite{roziere2023code}, and is among the most covered languages in open-source LLMs~\cite{gao2020pile,scao2022bloom} and network Q\&A platforms such as Stack Exchange. Furthermore, its syntax, for simple programs, resembles that of pseudo-code, thus abstracting from complex programming constructs. 

Metrics such as the accuracy, expressed as $\frac{1}{N}\sum_{i=1}^{N}\mathbbm{1} [\psi(y_i | x,p_i) = \Omega(p_i)]$, inform us as to the capabilities of a model.
Another metric we consider is how \emph{incorrect} a model is in its prediction, i.e., the average distance between the correct result and the model output, namely $\frac{1}{N}\sum_{i=1}^{N} |\psi(y_i | x,p_i) - \Omega(p_i)|$.
On tasks that require simulating multiple independent instructions or returning multiple correct predictions, we measure the prediction error as the Levenshtein distance between the prediction and the ground truth (as tuples), namely $\frac{1}{N}\sum_{i=1}^{N} |\psi(y_i | x,p_i) \cap \Omega(p_i)|$.
We also analysed the LLMs' responses to identify the most common reasons for failure in code simulation (and, by extension, task execution).

In summary, our methodology aims to reveal whether code simulation is a good proxy for naturalistic problems; we design reasoning tasks that can be turned into equivalent code, from operations such as addition and assignment to more complex constructs such as nested loops and sorting algorithms. 
We introduce five naturalistic and synthetic benchmarks: (I) \textbf{Straight line code simulation and Good exchange}, which tests the ability of an LLM to solve simple, sequential operations consistently; (II) \textbf{Critical path and Critical good exchange}, i.e., problems where only a portion is relevant to output the correct answer; (III) \textbf{Parallel path and Clique exchange}, which tests the capacity of an LLM to correctly perform multiple independent operations; (IV) \textbf{Nested loops and Recurring calculation}, that connects the computational complexity of a problem with the capabilities performing recurring reasoning operations, and (V) \textbf{Sorting and Ranking objects}, which tests sorting as a proxy of ranking objects with varying features.

\begin{figure*}
    \centering
    \includegraphics[width=1\linewidth]{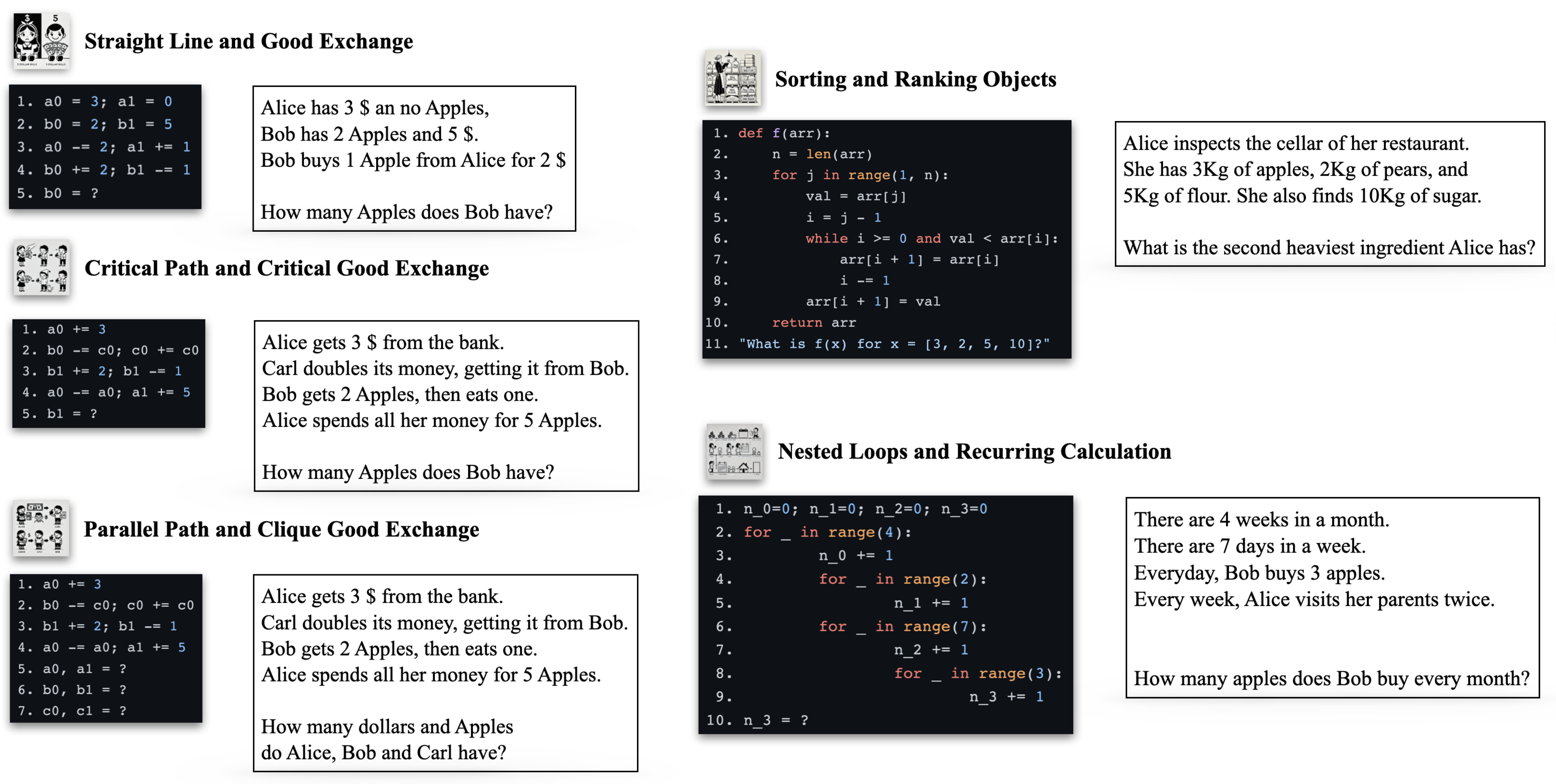}
    \caption{On the left, examples of Straight line and Good exchange tasks (top), Critical path and Critical good exchange (middle), and Parallel path and Clique good exchange (bottom). 
    On the right, examples of synthetic and naturalistic Sorting and Ranking objects (top) and Nested loops and Recurring calculation (bottom).}
    \label{fig:benchmarks-examples}
\end{figure*}
\subsection{Benchmarks}
In this section, we introduce the rationale behind each dataset. In particular, drawing from the literature in cognitive psychology~\cite{sweller1991evidence}, we focus on the reasoning capabilities that can be captured by purely synthetic tasks such as code.
Our framework benchmarks a model with pairs of synthetic and naturalistic prompts. If the correlation between the two settings holds strong, one can switch to purely synthetic benchmarks, for which generating new samples is much cheaper and scalable.
\paragraph{I. Straight line and Good exchange.}
Straight line code simulation can reveal whether a model processes sequential instructions faultlessly and leverages the principle of compositionality~\cite{dziri2023faith}. 
In this sense, Straight line can model scenarios that involve keeping track of objects whose state repeatedly changes over time~\cite{kim2023entitytrackinglanguagemodels}, as well as planning and Theory of Mind~\cite{kim2023fantombenchmarkstresstestingmachine,lin2024graphenhanced}. 
We model a naturalistic task as a Multi-Agent System, where different actors possess and exchange goods. Then, we ask a model to compute the number of goods (the value of a variable) an agent has after the exchanges.
We control the complexity of each problem by varying the number of operations/exchanges; other control variables can be introduced to study other capabilities, such as the number of agents and goods exchanged. 
The synthetic task that acts as a proxy of the naturalistic is a plain sequence of variable declarations, followed by instructions that modify the variables' state with sum and subtraction. 
Figure~\ref{fig:benchmarks-examples} (left) illustrates an example of the Straight line and Good exchange task.
In Appendix~\ref{a:straight-line}, we evaluate a richer set of operations, such as logical and/or, that is hard to model in naturalistic settings but can serve as a basis for increasing synthetic benchmarks. 
In addition to the Straight line task, we revisit the entity-tracking benchmark for LLMs~\cite{kim2023entitytrackinglanguagemodels} and pair it with a pure coding task where each object state's change is turned into a synthetic instruction with a variable assignment.

\paragraph{II. Critical path and Critical good exchange.}
Straight line and Good exchange do not assess a model's capabilities to distinguish relevant states from those that can be ignored. 
This setting is ubiquitous in machine learning and has been recently addressed in high-order tasks such as Theory of Mind~\cite{huang2024notion}.
We introduce a variation of the Straight line and Good exchange where only a fraction of the problem is sufficient to derive the correct answer. In this setting, agents exchange several goods. The critical path (i.e., the portion sufficient to solve the problem) is intertwined with other instructions that serve the role of noise or ``extraneous load'' in Cognitive Load Theory~\cite{sweller1991evidence}.
We synthetically model this setting with a well-known concept in algorithmic theory, i.e., that of a critical path, which is the stretch of dependent operations in a program.
Figure~\ref{fig:benchmarks-examples} (left) illustrates an example of the Critical path and Critical good exchange task.
Each good exchange is modelled as a variable addition, subtraction or assignment to resemble naturalistic operations. 
In Appendix~\ref{a:critical-path}, we evaluate a richer set of operations, such as logical and/or, that can serve as a basis to test increasingly complex naturalistic settings. 

\paragraph{III. Parallel paths and Clique good exchange.}
Another aspect of reasoning that complements the ``extraneous cognitive load'' is that of ``intrinsic load'' as the complexity of processing new information~\cite{sweller1991evidence}.
One can make a Straight line and a Good exchange problem more complex by requiring one to keep track of multiple variables/objects simultaneously. 
A model has to store, access, and modify all the variables consistently to return the correct result.
The Clique good exchange is a Multi-Agent Scenario that, paired with the Parallel paths, tests this setting.
Figure~\ref{fig:benchmarks-examples} (left) illustrates an example of such a setting. 
Similarly to the previous benchmarks, in Appendix~\ref{a:parallel-path}, we evaluate a richer set of operations, such as logical and/or, that are hard to model in naturalistic settings but can serve as the basis of increasingly complex synthetic benchmarks. 

\paragraph{IV. Nested loops and Recurring calculation.}
A set of problems often occurring in literature and human settings is recurring calculations, often in the form of math problems~\cite{cobbe2021trainingverifierssolvemath}.  
Consider this example: ``There are five working days a week. Every day of the week, I buy an apple. How many apples do I buy in 2 months?''. 
This and similar problems stress the capabilities of a model to connect the relevant pieces of information and compute recurring calculations, which can be further made more difficult by injecting noise, as per the previous example.
We design the Recurring calculation to test such capabilities. In code, nested loops with different indentations serve as a faithful proxy of naturalistic tasks with recurring calculations, as illustrated in Figure~\ref{fig:benchmarks-examples} (right).
On top of the experiments we conduct on pairs of synthetic and naturalistic prompts, interesting results emerge by solely analysing the capabilities of different models (such as Llama and GPT) on purely synthetic Nested loops, as reported in Appendix~\ref{a:nested-loops}.  

\paragraph{V. Sorting and Ranking objects.}
Ranking is a ubiquitous problem in both synthetic and naturalistic settings. To rank correctly, one must sort a set of objects by an attribute.
The Ranking objects problem prompts a model with a problem where an agent has to return the k$^{\text{th}}$-heaviest/lightest object in a group.
As Figure~\ref{fig:benchmarks-examples} (right) illustrates, the synthetic proxy problem consists of sorting a vector of numbers with an algorithm of choice, such as Bubble Sort or Insertion Sort.
This benchmark compares the reasoning capabilities of an LLM with synthetic and naturalistic sorting problems.\footnote{The optimal synthetic solution for the Objects ranking problem is a min-k heap, which is more efficient yet less intuitive than sorting. We thus focus on sorting.}
We also extensively evaluate sorting, i.e., pure code capabilities, with different algorithms of varying time- and space-complexity and show a tension between memorisation and code simulation capabilities in LLMs.
\begin{figure*}
    \centering
    \includegraphics[width=1\linewidth]{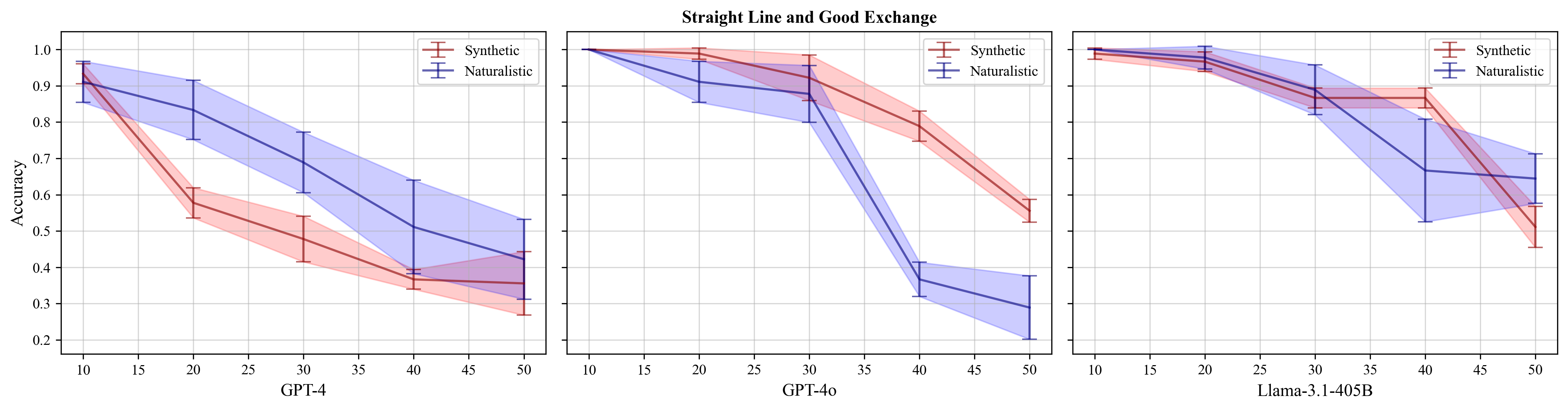}
    \includegraphics[width=1\linewidth]{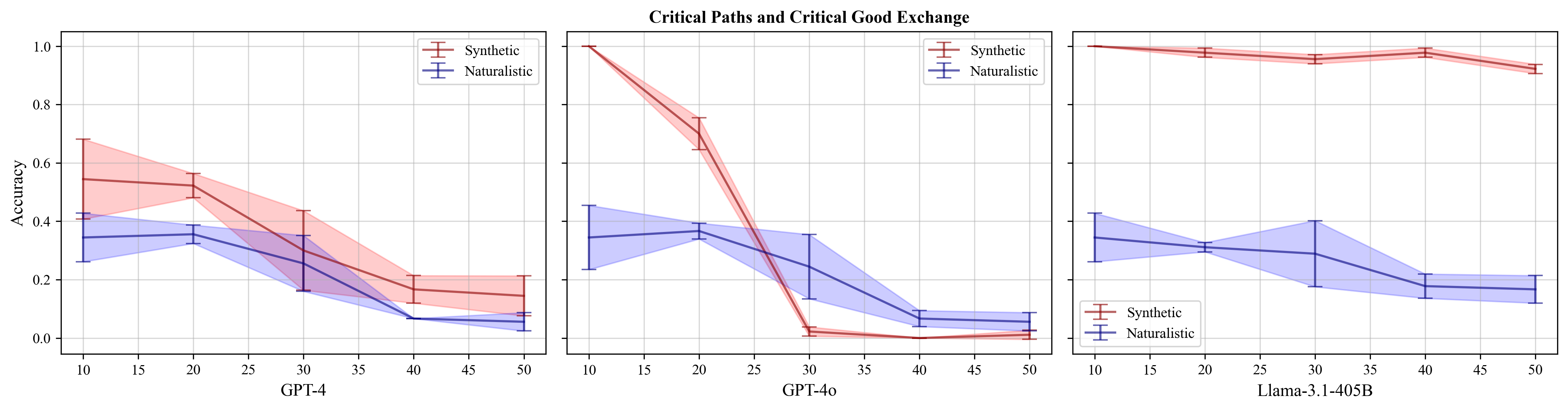}
    \includegraphics[width=1\linewidth]{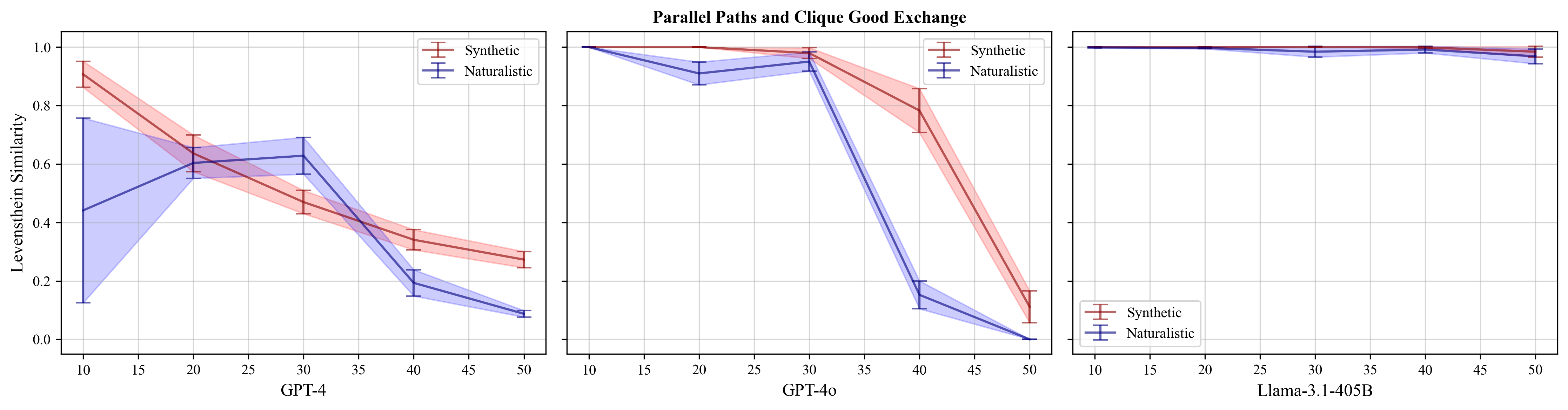}
    \caption{Top: Accuracy of different models on the Straight line and Good exchange.  
    Middle: Accuracy on the Critical path and Critical good exchange. The critical path length is 5.
    Bottom: Levenshtein similarity of the ground truth and the prediction on the Parallel path and Clique exchange. 
    The control variable for each problem is the number of operations, that spans from $10$ to $50$ with granularity $10$.}
    \label{fig:straight-line-results}
\end{figure*}

\section{Experimental Evaluation}
We conducted our experiments with open- and closed-source models, i.e., GPT-4, GPT-4o and Llama-3-405B. 
For each benchmark, we ran three independent runs with $30$ programs each to report the standard deviation over multiple trials. The control variable for the Straight line, Critical path, and Parallel path (and their naturalistic counterpart) is the number of instructions/exchanges. For Nested loops and Recurring operations, we vary the complexity of the task with the number of nested loops (i.e., the computational complexity of the algorithm to be simulated) or, equivalently, the number of nested recurring operations in the naturalistic task. For Sorting and Ranking objects, we vary the number of objects to rank.
Each run consists of naturalistic or synthetic prompts: we extract the answer and compare it to the ground truth label, obtained with a compiler/interpreter, to compute the performance metrics.
We evaluate each model with standard Chain of Thought~\cite{wei2022chain} (CoT). 
We also conduct an extensive evaluation of the synthetic capabilities of several LLMs (including GPT-3.5-Turbo, GPT-4, Llama-2, Llama-3, and Jurassic) on purely synthetic tasks, to highlight the reasons for failure when prompted to simulate code.
Finally, we emphasise that we only evaluate models without access to compilers/interpreters.

\subsection{Code Simulation as a Proxy of Naturalistic Tasks}
As reported in Figure~\ref{fig:straight-line-results} (top), the performance of Straight line code simulation matches that of the Good exchange, proving the synthetic task a faithful proxy of the performance of the naturalistic setting. Interestingly, GPT-4 is better on the naturalistic task while GPT-4o finds the synthetic task easier, with a ``gap'' between synthetic and naturalistic for high-complexity problems.
In Appendix~\ref{a:errors}, we conducted a linguistically informed analysis of the most frequent errors GPT-4o makes in the naturalistic setting. In the same section, we report the results and an analysis of the object tracking task introduced in~\cite{kim2023entitytrackinglanguagemodels}, which we pair with an equivalent coding task. We show that the two are very similar in performance, except for Llama-3.1-405B.
For the Critical path, Figure~\ref{fig:straight-line-results} (middle), we notice that the performance of GPT-4 strongly correlates with the synthetic and naturalistic tasks. Interestingly, GPT-4o and Llama-3.1-405B show similar trends for the naturalistic task, yet GPT-4o performs well for short programs, although its performance drops to zero as the program length increases. At the same time, Llama-3.1-405B always has better performance on the synthetic task. We conduct an informed linguistic analysis of the results and report it in Appendix~\ref{a:errors}.  
In the Parallel path and Clique good exchange (Figure~\ref{fig:straight-line-results}, bottom), all the models reveal a correlation between the naturalistic and synthetic settings. As we elaborate in Appendix~\ref{a:errors}, the gap of GPT-4o is caused by the model mishandling simple operations such as zeroing a variable or erroneously reusing the previous value of a variable.
\newline
Regarding Nested loops and Recurring operations, we notice that the performance of the naturalistic and synthetic tasks are correlated, yet the former is lower and affected by noise. 
We impute this behaviour to the intrinsic noise of natural language, and we confirm this with an in-depth linguistic analysis of the errors in Appendix~\ref{a:errors}.
While the correlation between the performances is still present, these results suggest that the naturalistic task is more challenging for LLMs.
Finally, for sorting, we notice ambivalent trends that suggest that each LLM handles the two tasks differently. Results reveal that the synthetic task becomes, for GPT-4 and GPT-4o, unexpectedly more straightforward for more challenging instances, while for Llama-3.1-405B, the behaviour of naturalistic and synthetic is monotonic and correlated. 
\newline
We dedicate the following subsections to providing a rationale for common failures of LLMs when they execute code.
We focus on the ``simulation gap'' caused by memorisation and shallow pattern recognition, revealing that LLMs can decide not to follow the prompt instructions while solving the task correctly, a phenomenon we name ``lazy execution regime''.
\begin{figure*}
    \centering
    \includegraphics[width=1\linewidth]{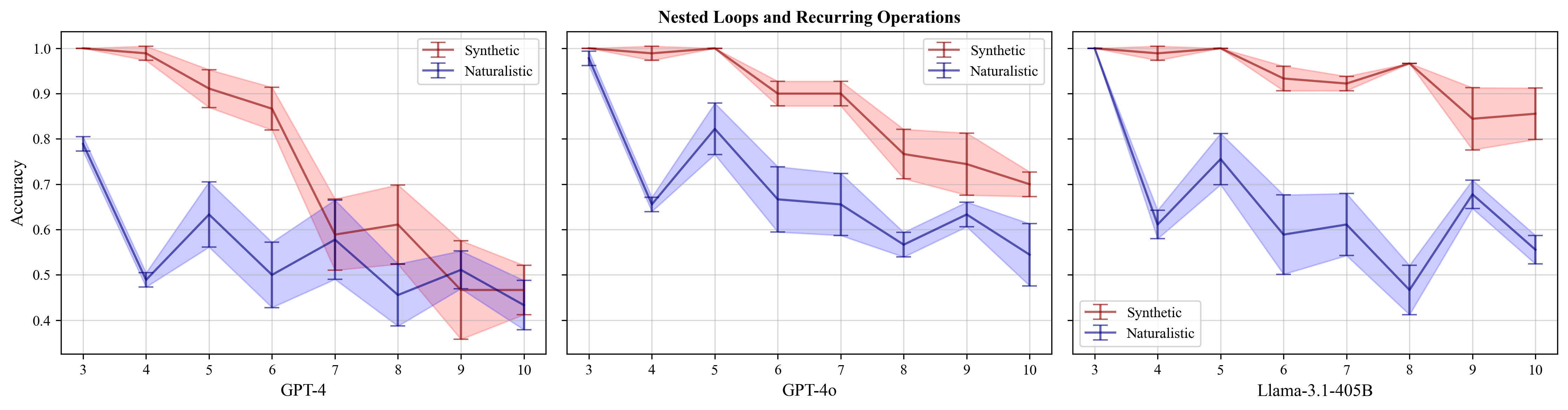}
    \includegraphics[width=1\linewidth]{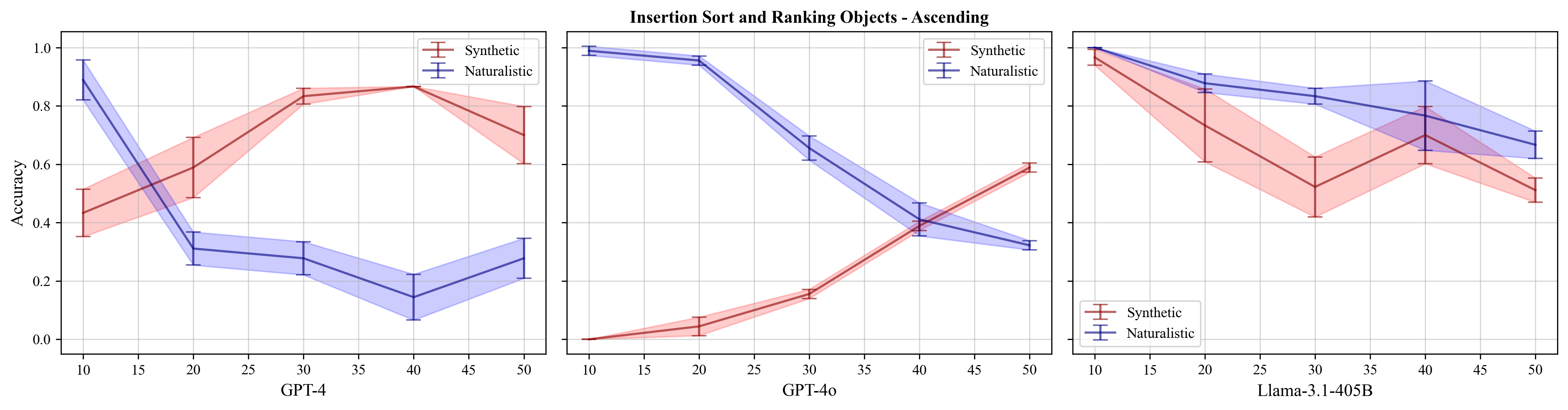}
    \includegraphics[width=1\linewidth]{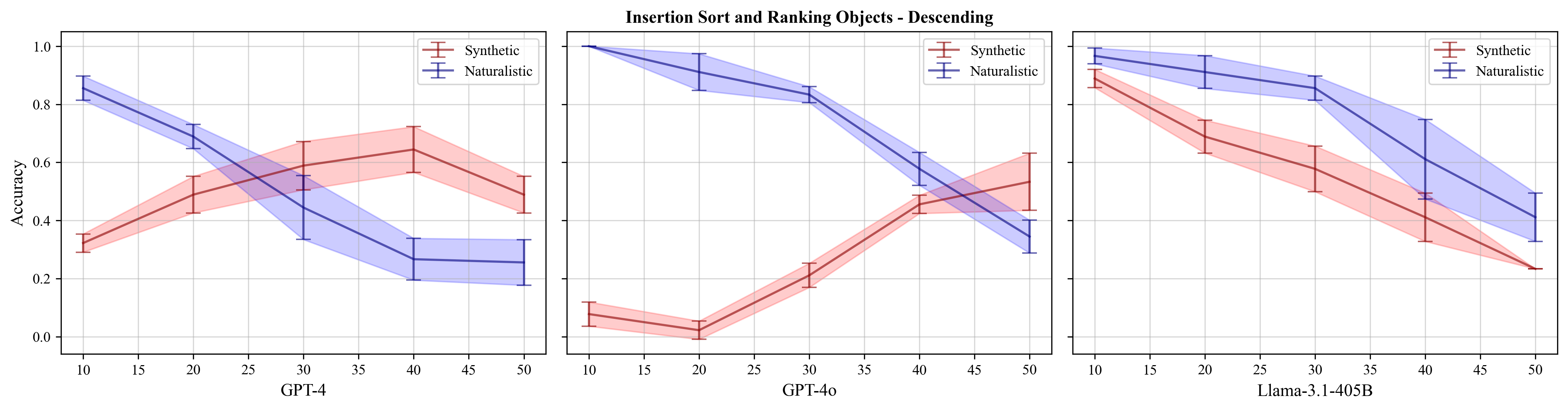}
    \caption{Top: Accuracy of different models on the Nested loops: the control variable is the max computational complexity of the task (synthetic) or, equivalently, the number of recurring operations (naturalistic). 
    Middle and bottom: Accuracy on the Sorting and Ranking objects task: the control variable is the number of elements to sort/rank to give the correct answer.}
    \label{fig:nested-sorting-results}
\end{figure*}

\subsection{The ``Simulation Gap'': Between Shortcuts and Memorisation}
\begin{figure*}
    \centering
    \includegraphics[width=1\linewidth]{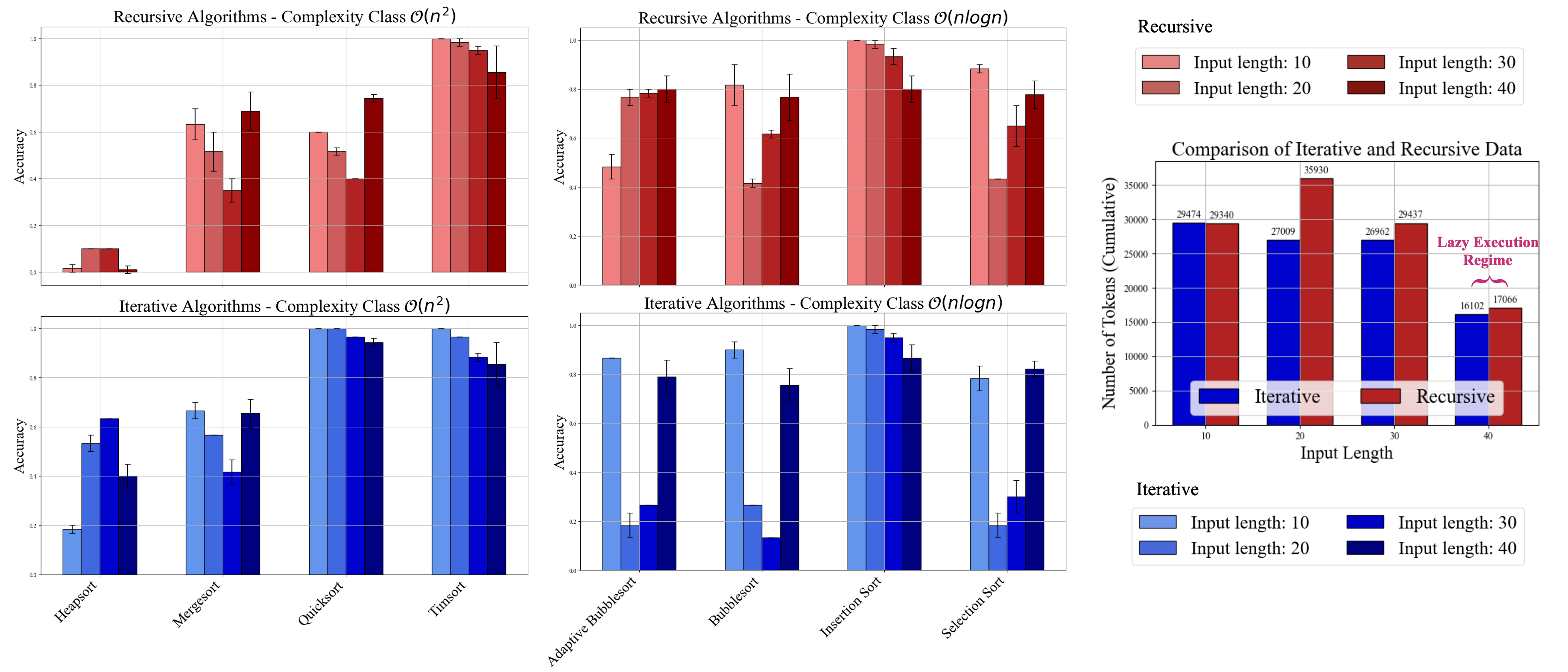}
    \caption{GPT-3.5-Turbo on sorting algorithms with varying complexity ($\mathcal{O}(n^2)$ and $\mathcal{O}(n \log n)$), both in their recursive (top) and iterative (bottom) versions. For long inputs, GPT-3.5 switches to a ``lazy execution regime'' (in \textbf{\textcolor{magenta}{magenta}}, right) where a model no longer simulates but just outputs the ordered sequence. Results on other models, including GPT-4, are reported in Appendix~\ref{a:sorting}.}
    \label{fig:sort-iterative}
    \vspace{-3mm}
\end{figure*}
\begin{figure}
\centering
    \includegraphics[width=1.\linewidth]{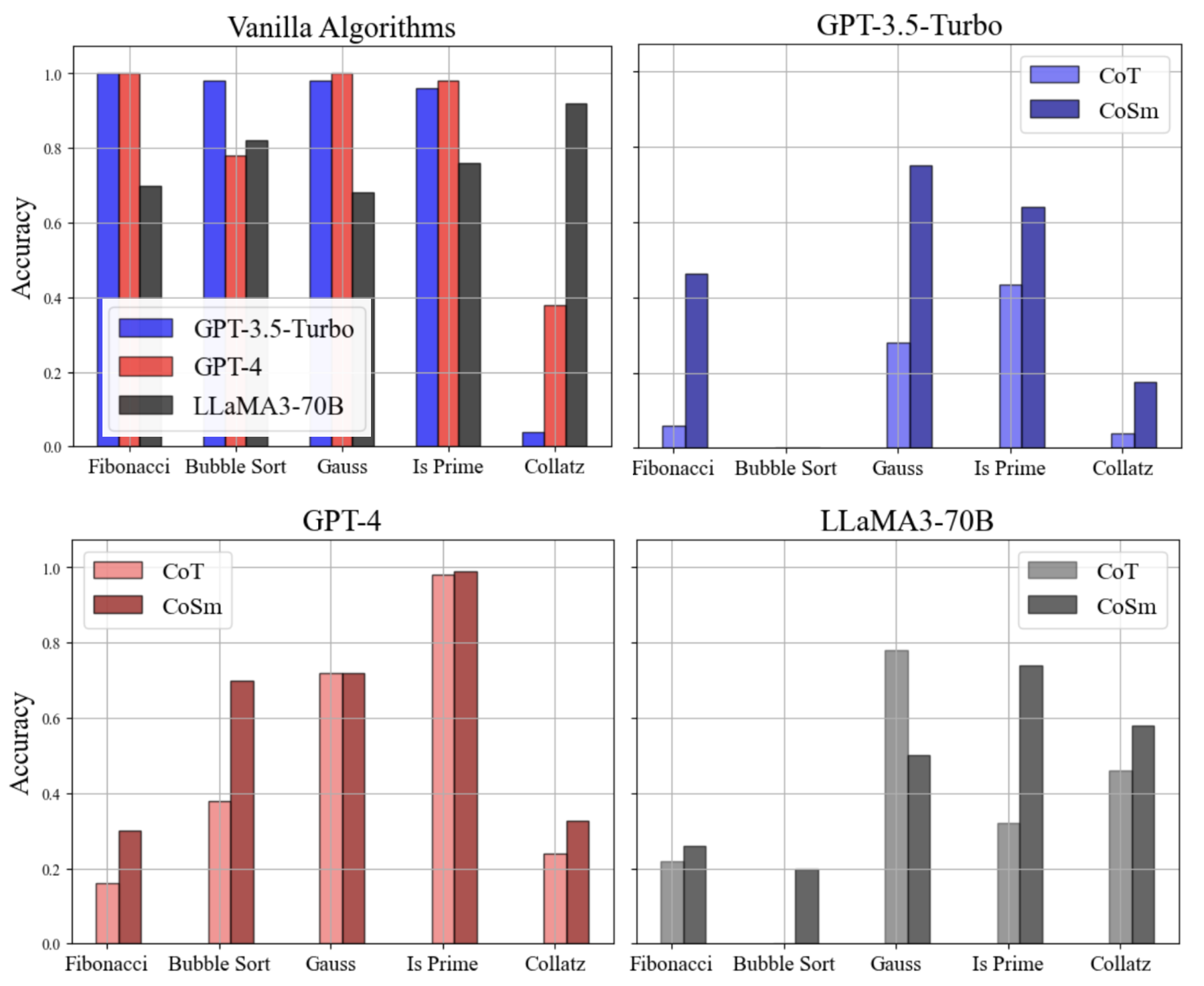}
    \caption{Results of GPT-3.5-Turbo, GPT-4 and Llama-3-70B on $50$ independent simulations of classic algorithms and their variations. Top-left: Performance on the \emph{vanilla} implementation of each algorithm with CoT. Top-right and bottom: Performance on the variations for each model with standard CoT vs. CoSm.}
    \label{fig:cosm}
\end{figure}
\begin{figure}
    \includegraphics[width=1.\linewidth]{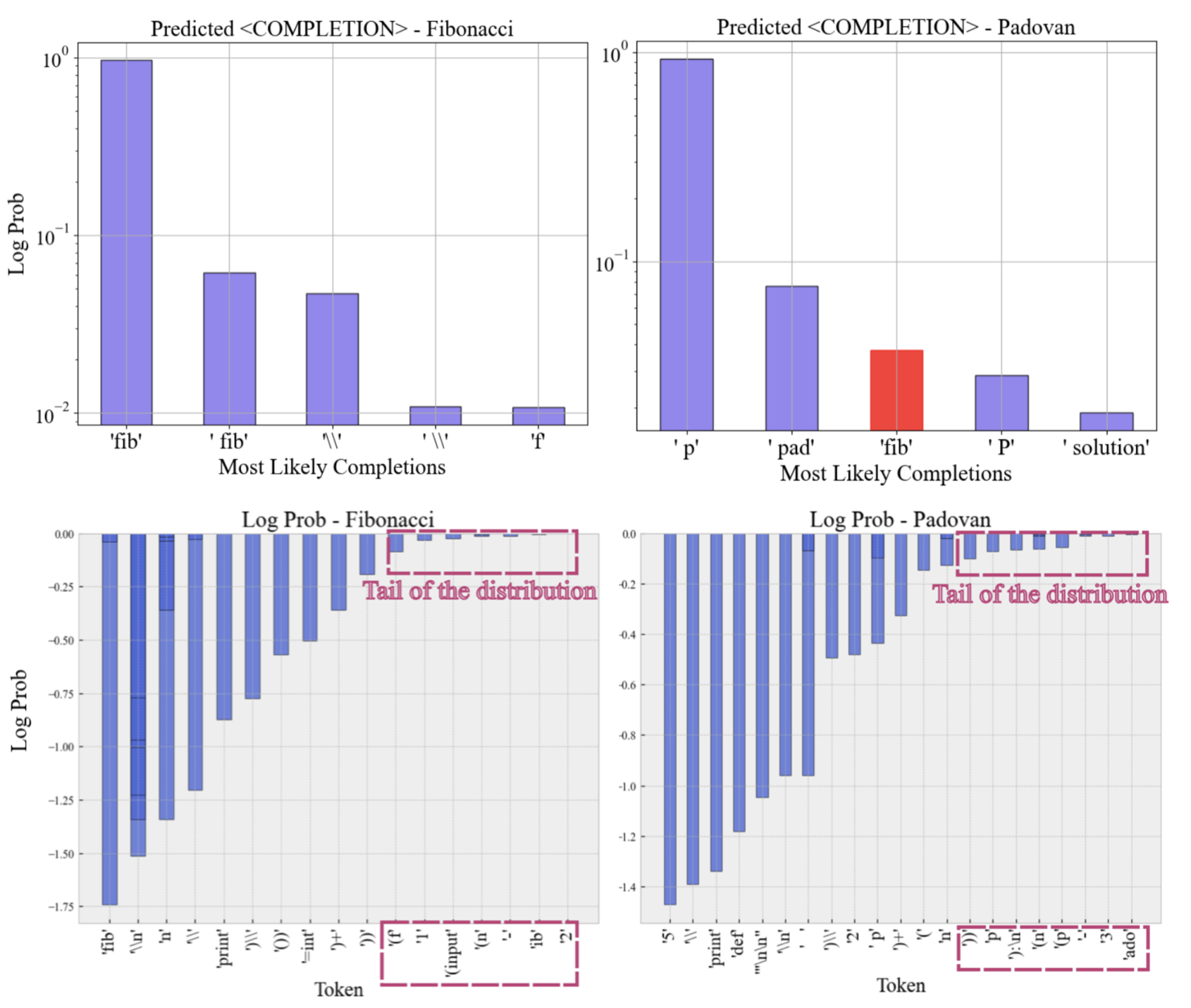}
    \caption{Top: Auto-completion of GPT-3.5-Turbo-Instruct when prompted to complete either the Fibonacci or Padovan function. 
    Bottom: A failure case of an anti-memorisation technique~\protect\cite{shi2023detecting}: the least likely predicted tokens of both Fibonacci and Padovan overlap and are not informative of memorisation.}
    \label{fig:barplot-fibo-pado}
    \vspace{-5mm}
\end{figure}
\paragraph{A ``lazy'' code simulation regime.}
To deepen our understanding of why LLMs fail on sorting routines, we assessed whether they can simulate 8 sorting routines in their iterative and recursive versions and with log-linear or quadratic complexity. 
Each input is a vector of varying length ($\{10, 20, 30, 40\}$), where each element is an integer randomly sampled between $0$ and $100$. All the routines are reported in Appendix~\ref{a:sorting-code}.

We study GPT-3.5-Turbo, GPT-4, and Llama-3-70B and run $3$ independent runs of $30$ experiments each are shown in Figure~\ref{fig:sort-iterative}. We analyse GPT-3.5-Turbo here as it results in the most interesting findings, though the same trends affect GPT-4 and Llama-3; see Appendix~\ref{a:sorting}.
Beyond confirming the implicit bias of Transformers towards ordered sequences~\cite{dufter2022position} (i.e., they tend to output ordered sequences), Figure~\ref{fig:sort-iterative} (left) shows that LLMs often provide the correct result for classic sorting algorithms such as Insertion Sort. However, with less common algorithms, LLMs trace the expected execution but fail even for vectors of a few inputs.
In summary, we find that (i) standard sorting algorithms (e.g., Insertion and Quick Sort) lead to more accurate results; (ii) there is a weak correlation between algorithm complexity and the accuracy of its simulation, thus reinforcing the hypothesis that in this case, the model is not simulating the procedure; (iii) there is a weak correlation between the length of the input vector and the accuracy of a model. 

By contrast, GPT-3.5-Turbo is accurate on input vectors of max length (i.e., $40$). 
We name this phenomenon ``lazy execution regime'' (in \textbf{\textcolor{magenta}{magenta}}, Figure~\ref{fig:sort-iterative}, right): the number of tokens generated for long sequences is considerably smaller than for shorter inputs, hinting that the task's induction bias is more evident when the input is a consistent portion of the prompt. 
Finally, we document an endemic case of failure with LLMs such as GPT-3.5-Turbo and GPT-4. On standard algorithms such as Bubble Sort, long input sequences often lead to the wrong result when they contain repeated elements, as we document in Appendix~\ref{a:sorting}. We hypothesise that the probability of a sequence that contains repeated elements is so low, conditioned on what has been generated so far, that a model skips repeating elements, even when we set the `presence penalty' value to zero.

\paragraph{The pitfalls of memorisation with code.}
As the main research question of this work is whether one can use code as a proxy of high-order/reasoning tasks, it is important to investigate the role of memorisation on common algorithms, as they may bias the comparison to naturalistic, non-memorised tasks. We paired five standard sorting algorithms with slight variations that neither affect the code length nor their computational complexity. Yet, their semantics are slightly changed to produce a different output. We also introduce a prompting technique that, on top of CoT prompting, explicitly instructs a model to simulate a routine step by step: we name this prompting method Chain of Simulation, or CoSm (the prompt we use is reported in Appendix~\ref{a:cosm}).
We investigated the following algorithms and their variations: (i) the \textbf{Fibonacci} sequence paired with \textbf{Padovan}, a slight variation that modifies the return condition; (ii) \textbf{ascending Bubble Sort}, paired with the \textbf{descending} routine; (iii) the \textbf{Gauss algorithm}, paired with a variation that, instead of summing the first $n$ natural numbers, \textbf{adds even and subtracts odd numbers}; (iv) a \textbf{primality test} paired with the same routine on the \textbf{successor} of the input; (v) and the sum of the first $n$ \textbf{Collatz numbers}, paired with a variation that returns the sum of the \textbf{even numbers} in a Collatz sequence. 
All the functions and their variations were first anonymised to avoid bias towards known function names and implemented with the code that appears most frequently on GitHub. We report their implementation in Appendix~\ref{a:classic-algs}.
As shown in Figure~\ref{fig:cosm} (top), GPT-3.5-Turbo, GPT-4 and Llama-3-70B are accurate on each classic algorithm, but their accuracy dropped significantly with the variations. On the other hand, their accuracy is marginally improved when we prompt them explicitly to simulate the routine with CoSm (Figure~\ref{fig:cosm}, bottom).
We observe that LLMs anticipate the behaviour of a function by looking at specific dominating patterns. 
As shown in Figure~\ref{fig:barplot-fibo-pado} (right), GPT-3.5-Instruct completes a template of the Fibonacci function with tokens compatible with the function's scope. On Padovan, there is a non-negligible probability (the third most likely token, in \textbf{\textcolor{red}{red}}) that the predicted function is Fibonacci.
These results question the induction head mechanism~\cite{olsson2022context}, as a slight variation of a task where an LLM is accurate results in errors of a large order of magnitude. 
Furthermore, this particular type of uncertainty goes unnoticed with methods that detect memorisation~\cite{shi2023detecting}, as shown in Figure~\ref{fig:barplot-fibo-pado} (right). While anti-memorisation techniques are effective with textual inputs (e.g., the Wikipedia dataset), they fail on code. We hypothesise that low-frequency tokens are informative for natural language but not on code as it is more structured, with low-likelihood tokens (in \textbf{\textcolor{magenta}{magenta}}) that are not predictive of a model's memorisation.

\section{Conclusions and Future Work}
In this work, we introduce a way to evaluate some high-order capabilities of LLMs via synthetic tasks in the form of code.
We show that many reasoning tasks have a natural reformulation as code, which is easy to obtain and scale.
Our experiments show the feasibility of our approach and pave the way to synthetic benchmarks: on the other hand, code simulation may suffer from memorisation (for common routines such as sorting), and one has to carefully consider these drawbacks when designing a synthetic task as a proxy of high-order reasoning.

\section*{Acknowledgments}
ELM is supported by the Alan Turing Institute. AGC is supported by the Economic and Social Research Council (ESRC) under grant ES/W003473/1, by the Fundamental Research priority area of The Alan Turing Institute, and by the Turing Defence and Security programme, through a partnership with the UK government in accordance with the framework agreement between GCHQ and the Alan Turing Institute. 
We thank Microsoft Research - Accelerating Foundation Models Research program for the provision of Azure resources to access OpenAI models.

\clearpage
\bibliography{example_paper}
\bibliographystyle{icml2025}

\newpage
\appendix
\onecolumn

\section{The Simulation Gap: Further Analyses of Naturalistic and Synthetic Tasks}~\label{a:errors}
\begin{figure*}[!ht]
    \centering
    \includegraphics[width=1\linewidth]{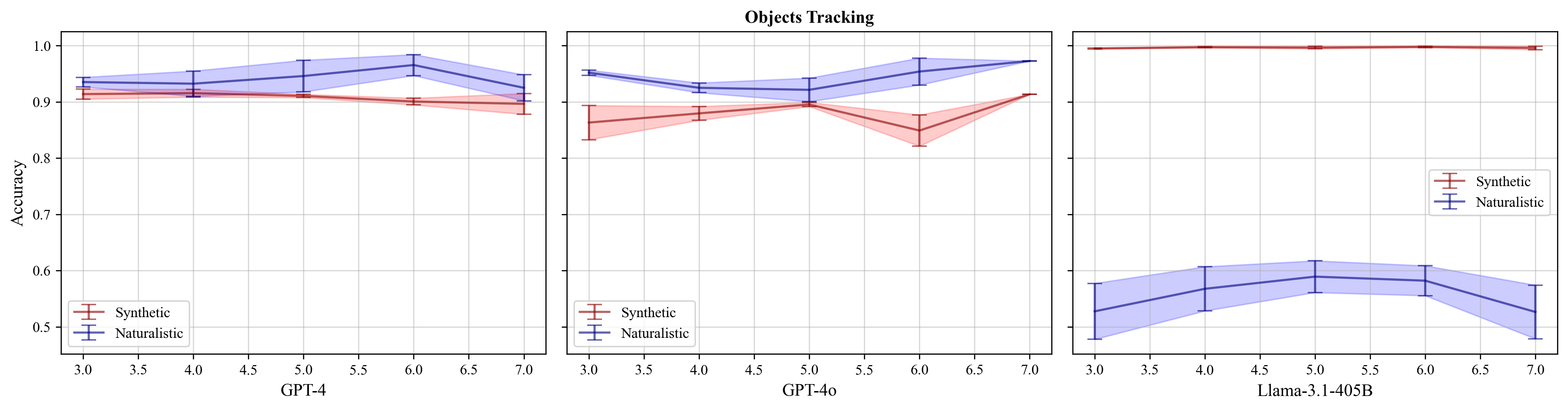}
    \caption{Accuracy of different models on the Object tracking task introduced in~\cite{kim2023entitytrackinglanguagemodels}. 
    We pair the naturalistic task with an equivalent coding task.}
    \label{fig:kim-schuster}
\end{figure*}
\subsection{Straight Line and Good Exchange}\label{a:straight-line-analysis}
\paragraph{A linguistic analysis of the simulation gap.} Across the incorrect solutions, which can be found in the code material and in particular in \texttt{`logs/straight-line/n\_ops-40\_n\_vars-3\_n\_instances-2\_batch-1.json'}, there are recurring issues like incomplete step-by-step tracking—where the model misinterprets large quantities or forgets to zero out a sender’s items or incorrectly transferring the full amount to the recipient. 
The model also often updates only one side of an exchange (like subtracting from one agent but not adding to another), merges or ignores consecutive actions (for example, combining multiple ``buy'' actions), or makes arithmetic slips that throw off subsequent counts. Sometimes, it simply resets to a wrong intermediate value or cuts off its reasoning prematurely and provides a final answer without incorporating all of the steps.

\begin{figure*}
    \centering
    \includegraphics[width=1\linewidth]{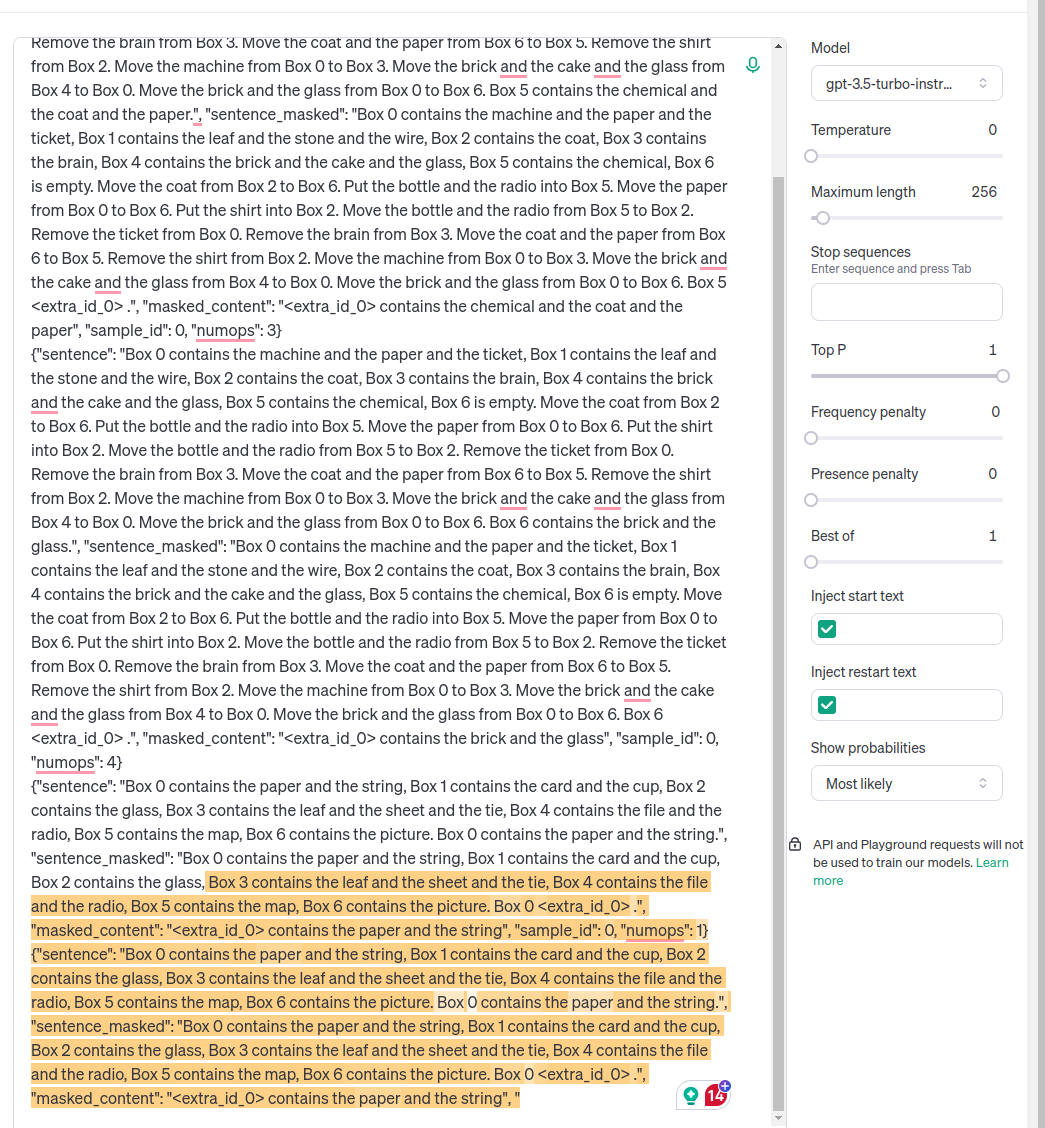}
    \caption{Memorisation of the Object tracking dataset~\cite{kim2023entitytrackinglanguagemodels}. When GPT-3.5-Turbo-Instruct is fed with some inputs from the dataset, it can recover the continuation almost verbatim (highlighted in yellow), a hint the model has memorised the dataset. For this example, we used the OpenAI Playground.}
    \label{fig:kim-schuster-memo}
\end{figure*}

\paragraph{Object tracking~\cite{kim2023entitytrackinglanguagemodels}.}
As reported in Figure~\ref{fig:kim-schuster}, GPT-4 and GPT-4o have very good performance on both the synthetic and naturalistic tasks, while there is a strong gap in favour of the synthetic dataset for Llama-3.1-405B.
We run a memorisation test on the naturalistic dataset, which has been well studied in many works and released two years ago. By feeding some truncated inputs of the naturalistic dataset to GPT-3.5-Turbo-Instruct, we were able to recover the remaining part of the input and the label, a strong hint that the dataset has been memorised verbatim by GPT models. We report a test of memorisation in Figure~\ref{fig:kim-schuster-memo}.

\subsection{Critical Path and Critical Good Exchange and Parallel Path and Clique Good Exchange}\label{a:critical-path-analysis}
Across these incorrect solutions, which can be found in the code material and in particular in which can be found in the code material and in particular in \texttt{`logs/critical-path/n\_ops-30\_n\_vars-6\_len\_critical\_path-5\_batch-3.json'} (the same errors also affect Parallel Path and Clique Good Exchange) the model, frequently truncates its step-by-step reasoning midway, skips the update for certain variables, and makes arithmetic errors involving negative signs or repeated multiplication (by two). It also struggles with handling variables after they are reset to zero, sometimes reusing outdated values or forgetting that the variable is now zero. In many instances, the model fails to complete all lines of code when attempting to explain or execute them, likely due to internal length constraints or confusion as it walks through numerous instructions. As a result, its chain of thought becomes inconsistent, causing final answers to be incorrect or absent altogether. These errors are especially pronounced in problems requiring precise step-by-step arithmetic and frequent reassignment of variables, where any small oversight can render the final result invalid.

\clearpage

\section{An Analysis of the Code Simulation Capabilities of LLMs}
\paragraph{Replicability.} The code to replicate the experiments on pure code simulation is available \href{https://github.com/EmanueleLM/CodeSimulation}{here}.

\subsection{Straight Line}\label{a:straight-line}
This section describes our results with straight-line programs, code with critical paths, and approximate and fault-tolerant instructions.
We then study nested loops and sorting algorithms. The key controlled variable for the input is the number of instructions, in line with recent works in the area~\cite{zhou2023algorithms}. 

\subsection{Straight-line Programs Simulation}
\begin{wrapfigure}{R}{0.3\textwidth}
\vspace{-5mm}
\begin{minipage}{0.3\textwidth}
\begin{lstlisting}
 a0=-1; a1=0; a2=-6
 a1 += a2
 a0 = a2
 a0 -= a0
 a1 = a0
 a0 -= a2
\end{lstlisting}
\vspace{-2mm}
\caption{Straight-line code.}\label{code:straight-line}
\end{minipage}
\vspace{-2mm}
\end{wrapfigure}
\begin{figure*}
\centering
\includegraphics[width=0.8\linewidth]{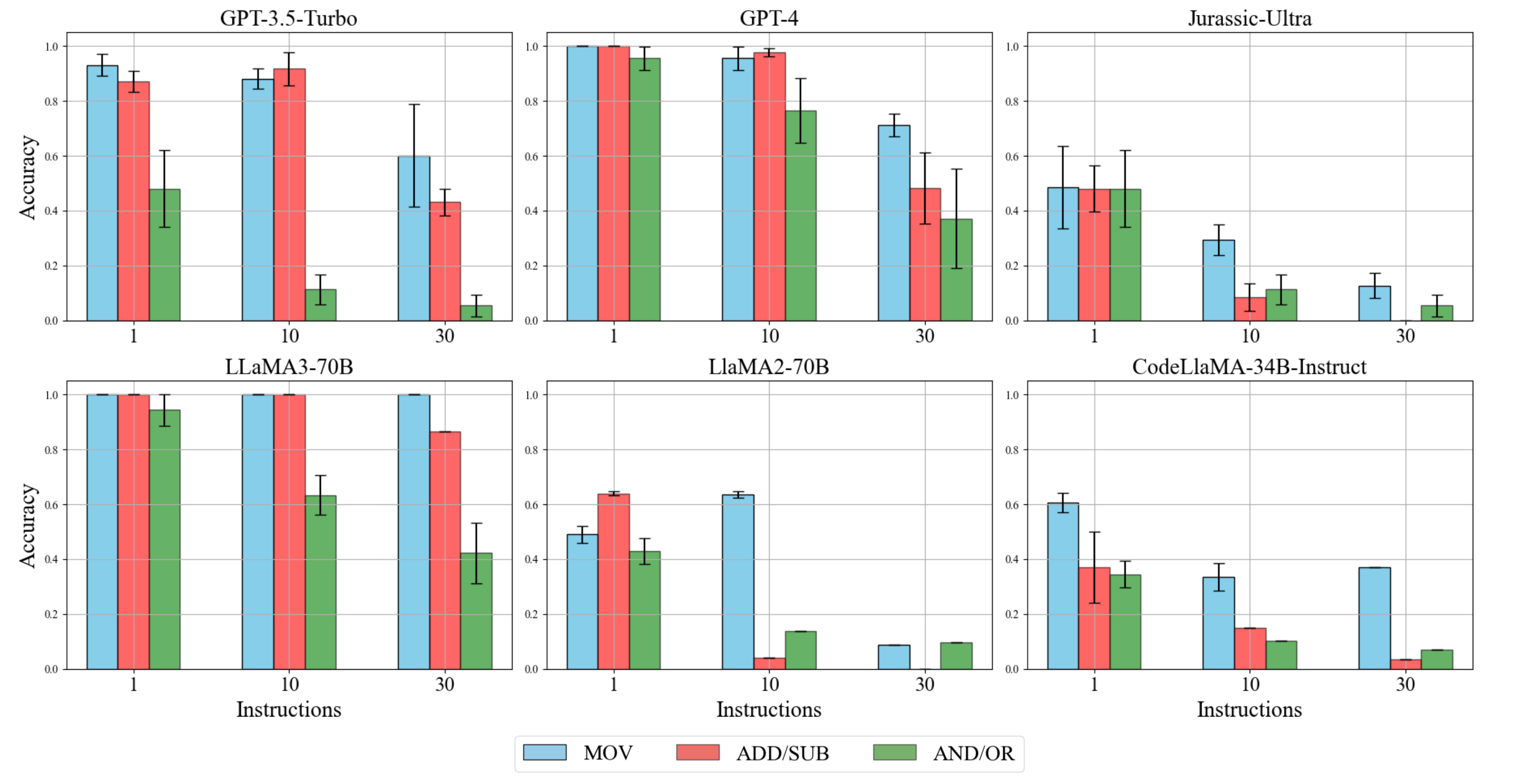}
\caption{Accuracy 
on $3$ independent runs of $30$ experiments each of different LLMs on code snippets with \textbf{solely} \{\texttt{and, or}\}, \{\texttt{add, sub}\} or \{\texttt{mov}\} instructions. We group results by codes of varying number of instructions (x-axis), namely $\{1, 10, 30\}$.}
\label{fig:llm-cpu-mixed-instructions}
\end{figure*}
\begin{figure*}
\centering
\includegraphics[width=\linewidth]{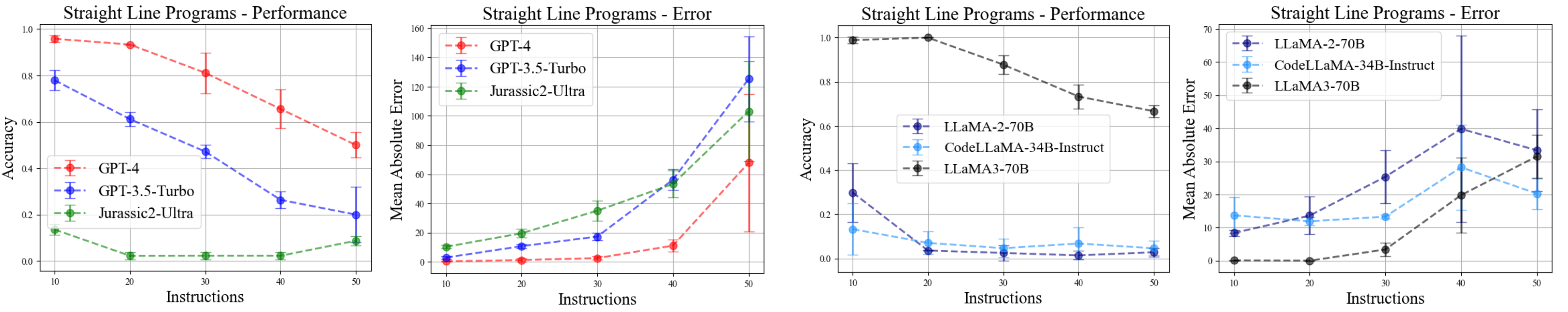}
\caption{Accuracy and Mean Absolute Error of different LLMs on code of varying length with \textbf{only} \{\texttt{add,sub}\} and \{\texttt{mov}\} instructions (out of $3$ independent runs of $30$ experiments each).}
\label{fig:llm-cpu-single-instructions}
\vspace{-4mm}
\end{figure*}

We first assess the simulation capabilities of different LLMs on code that contains only \{\texttt{add,sub}\}, \{\texttt{mov}\}, or logical-\{\texttt{and,or}\} instructions.
Figure~\ref{fig:llm-cpu-mixed-instructions} shows that for code containing only one type of instruction, Jurassic2-Ultra, Llama-2-70B and CodeLLaMA-34b-Instruct are poor code simulators: their performance significantly downgrades with just $10$ instructions, while GPT-3.5-Turbo, GPT-4 and Llama-3-70B are more accurate, though the same detrimental effect is evident, for example, on programs with $30$ sequential instructions. 
Logical instructions are hard to simulate for any model (green bar): we hypothesise that the reason is their low coverage in the training set since even a simple neural network can correctly compute logical-\{\texttt{and,or}\}.
Since the performance of any model considerably drops with logical-\{\texttt{and,or}\} instructions, we exclude such operation and synthesise straight-line programs with $\{10, 20, 30, 40, 50\}$ lines of instructions and a fixed number of variables (e.g., 5), as shown in Figure~\ref{code:straight-line}. We then prompt an LLM to compute the value of one of such variables at the end of the execution.
Both settings prompt each LLM to predict the state of a variable at the end of the computation.
Figure~\ref{fig:llm-cpu-single-instructions} shows our results for code with mixed instructions: in this task, they successfully achieve compositionality and reliably simulate code with mixed instructions.  GPT-4 and Llama-3 are reliable instruction simulators, followed by GPT-3.5-Turbo. Conversely, Jurassic2-Ultra, Llama-2-70B and CodeLLaMA-34B cannot simulate even short snippets of instructions.
Qualitatively, we further note that most errors occur when the output of the LLM consists only of the final result of the computation rather than the complete program execution trace.\footnote{We observed this phenomenon with GPT-4 in more than $95\%$ of cases, as per the code attachment.}

\subsection{Critical Path}\label{a:critical-path}
\begin{wrapfigure}{R}{0.35\textwidth}
\vspace{-5mm}
\begin{minipage}{0.35\textwidth}
\begin{lstlisting}[escapechar=\%]
 a0 = a1 = a2 = 1
 %\textcolor{red}{a3 = a4 = a5 = -1}%
 a0 -= a1
 %\textcolor{red}{a3 -= a4}%
 %\textcolor{red}{a5 \&= a3}%
 %\textcolor{red}{a3 |= a5}%
 a0 += a1
 a1 -= a3
\end{lstlisting}
\vspace{-2mm}
\caption{Code with \textcolor{red}{critical path}. 
}\label{code:critical-path}
\end{minipage}
\vspace{-3mm}
\end{wrapfigure}
Some sequential problems can be solved without executing all the instructions in a program. For instance, consider the code in Figure~\ref{code:critical-path}, with a model prompted to predict the value of \verb~a3~.
To compute the value of \verb~a3~, it suffices to execute only those code blocks highlighted in \textcolor{red}{red}, which we refer to as the \textcolor{red}{critical path} of \verb~a3~.\footnote{From a theoretical perspective, a neural network that isolates a variable's critical path is straightforward to build, as shown in the Appendix, Section~\ref{a:critical-path}.}
We thus perform experiments with programs that contain critical paths shorter than the entire program.

\begin{figure*}
\centering
\includegraphics[width=\linewidth]{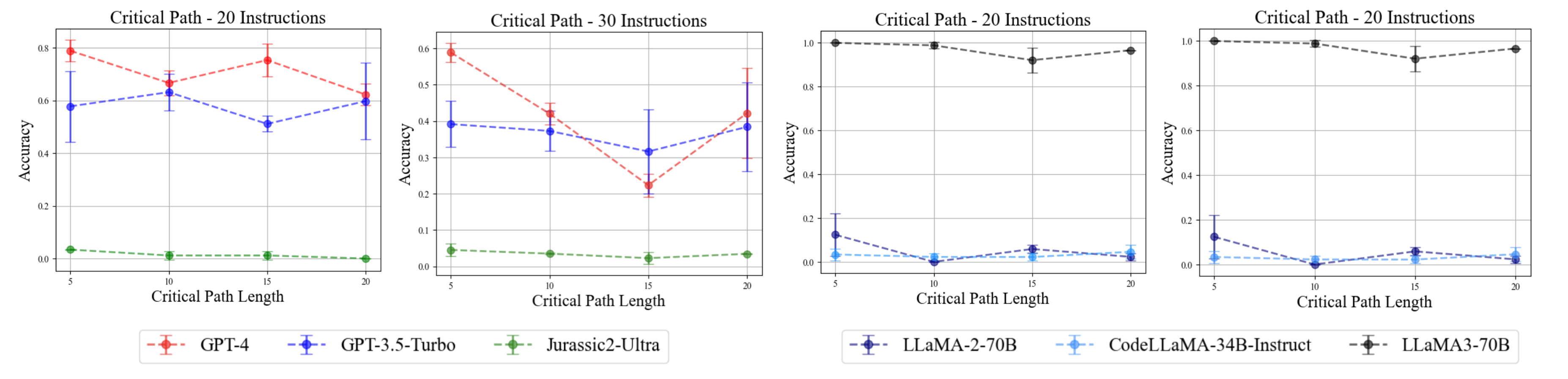}
\caption{Accuracy of different LLMs on $3$ runs of $30$ experiments each on programs with varying critical path lengths, for snippets of $20$ and $30$ lines of code respectively.}
\label{fig:code-complexity-critical-path}
\vspace{-5mm}
\end{figure*}

In Figure~\ref{fig:code-complexity-critical-path}, and for $3$ independent runs with $30$ programs each, we present the results when GPT-3.5-Turbo, GPT-4, Jurassic2-Ultra, Llama-3-70B, Llama-2-70B and CodeLLaMA-34b-Instruct are prompted to execute snippets of $20$ and $30$ instructions, with critical paths of varying length (i.e., $\{5, 10, 15, 20\}$).
GPT-4 and Llama-3-70B can leverage smart execution, though Llama-3-70B is better than GPT-4, especially on $30$ lines of code.
Although GPT-4's general simulation accuracy is higher than GPT-3.5-Turbo, it is less robust to variations of critical path length, i.e., GPT-4 suffers from a more severe accuracy drop compared to GPT-3.5-Turbo when critical path length approaches that of the entire program.
We also notice that Llama-2-70B, CodeLLaMA-34B and Jurassic2-Ultra cannot generally execute instructions reliably.
As with straight-line execution, we notice that most errors occur when the output trace contains only the result, not the code simulation.

\subsection{Parallel Path}\label{a:parallel-path}
\begin{figure}
\centering
\includegraphics[width=1\linewidth]{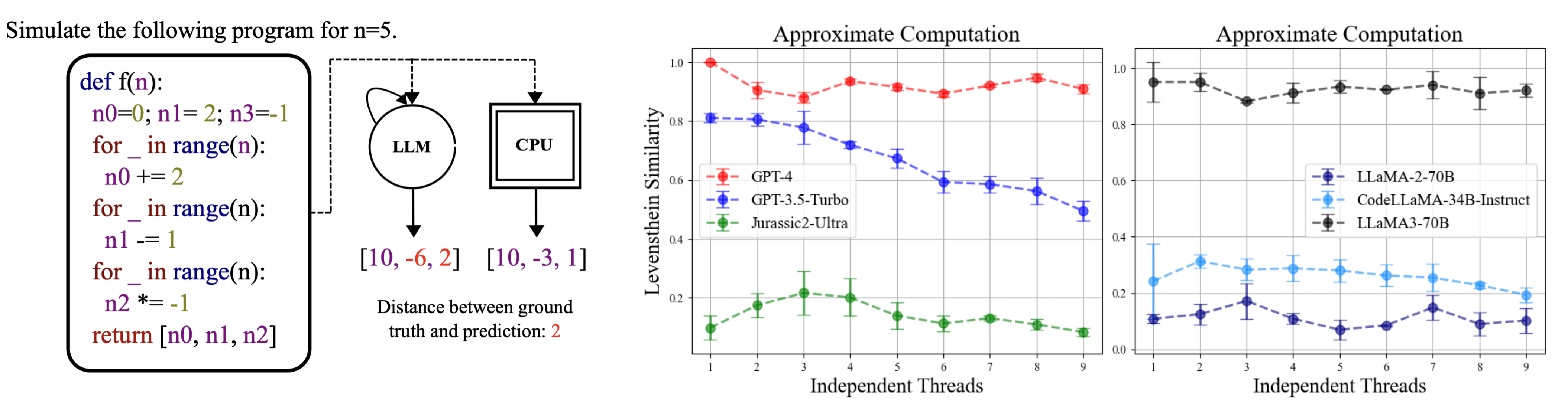}
\caption{On the left, an example of an algorithm to test the approximation capabilities of a model. On the right, the Levenshtein similarity between the ground truth and an LLM's output measures the performance of different models (the higher, the better).}
\label{fig:approximate-example}
\vspace{-5mm}
\end{figure} 
\textbf{Approximate computation} is evaluated with programs of $k$ \texttt{for} loops with $n$ instructions each. Each loop independently contributes to the final function return value, as shown in Figure~\ref{fig:approximate-example}. 
We denote by $\delta$ the probability of wrongly computing the result of each independent loop so that the probability of computing the exact result for a consistent \emph{analog} computer on a program is $(1-\delta)^k$
. For an LLM, $\delta$ is computed as the Levenshtein similarity between the ground truth values and the predicted results and is a proxy for the approximation capabilities on programs of varying complexity.
Results are reported in Figure~\ref{fig:approximate-example}.
GPT-4 and Llama-3 are the best-performing models, with no accuracy degradation even for long programs with up to nine independent \emph{threads}.

\begin{figure}
\centering
\includegraphics[width=1\linewidth]{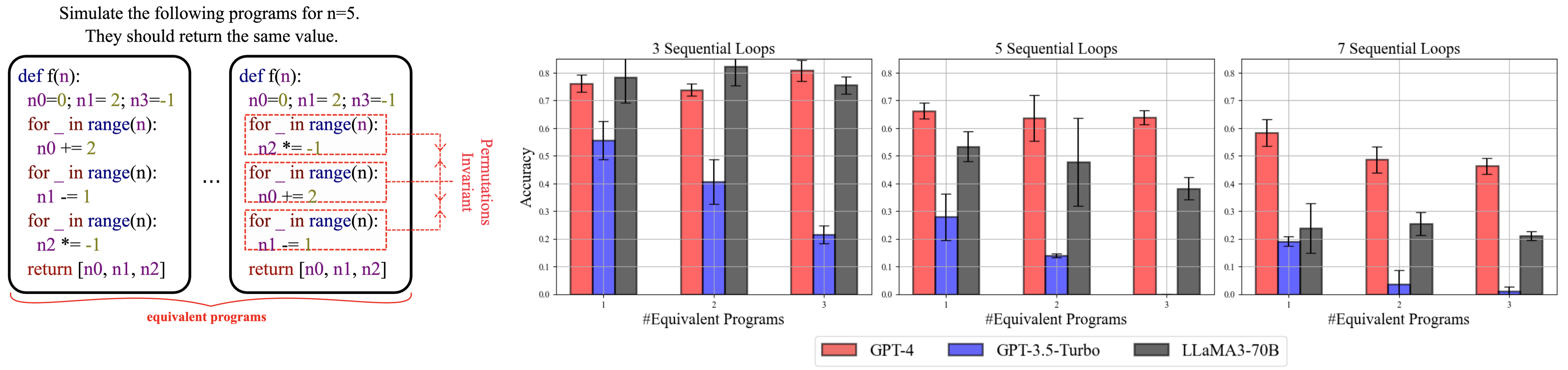}
\caption{Left: an example of a fault-tolerant algorithm. We feed an LLM with a few equivalent programs and instruct it to execute all of them to return the same result.
Right: how redundancy affects the performances of GPT-3.5-Turbo, GPT-4 and Llama-3-70B on multiple equivalent programs.}
\label{fig:fault-tolerant}
\vspace{-5mm}
\end{figure}

A routine is \textbf{tolerant to faults} when it can recover from errors occurring during the computation. 
To test LLMs in this setting, we prompt a model with different variations of the same algorithm, specifying that the objective is to demonstrate they yield the same result.
Figure~\ref{fig:fault-tolerant} (left) reports an example of fault-tolerant prompts. The illustration is complemented by results for different LLMs on three independent runs of $30$ experiments each; the control variable is the number of equivalent programs fed to the model.
While redundancy neither alters GPT-4 accuracy nor improves its performance, GPT-3.5-Turbo is heavily affected by multiple equivalent programs in the prompt and experiences a severe decrease in performance. Results for Jurassic2-Ultra, Llama-2-70B and CodeLLaMA-34B evidence low accuracy and are excluded from the evaluation (though inspectable in the code).

\subsection{Nested Loops}\label{a:nested-loops}
\begin{wrapfigure}{R}{0.38\textwidth}
\begin{minipage}{0.38\textwidth}
\centering
\vspace{-2mm}
\includegraphics[width=1\linewidth]{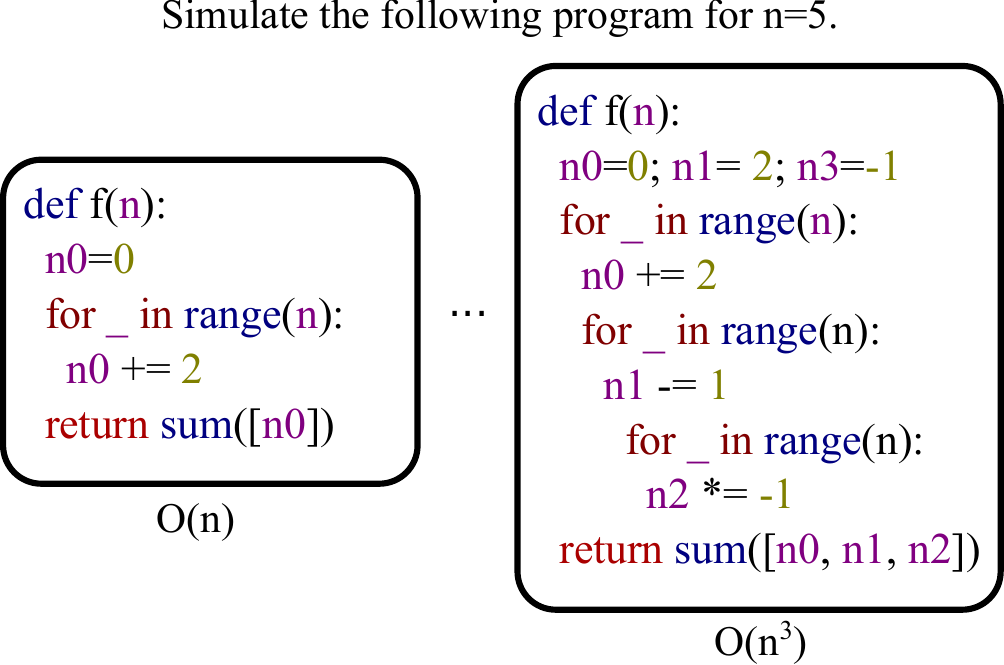}
\caption{Examples of programs with varying computational complexity: on the left, linear ($\mathcal{O}(n)$), on the right, cubic ($\mathcal{O}(n^3)$).}
\label{fig:nested-loops}
\end{minipage}
\vspace{-3mm}
\end{wrapfigure}
\begin{figure*}
\centering
\includegraphics[width=1\linewidth]{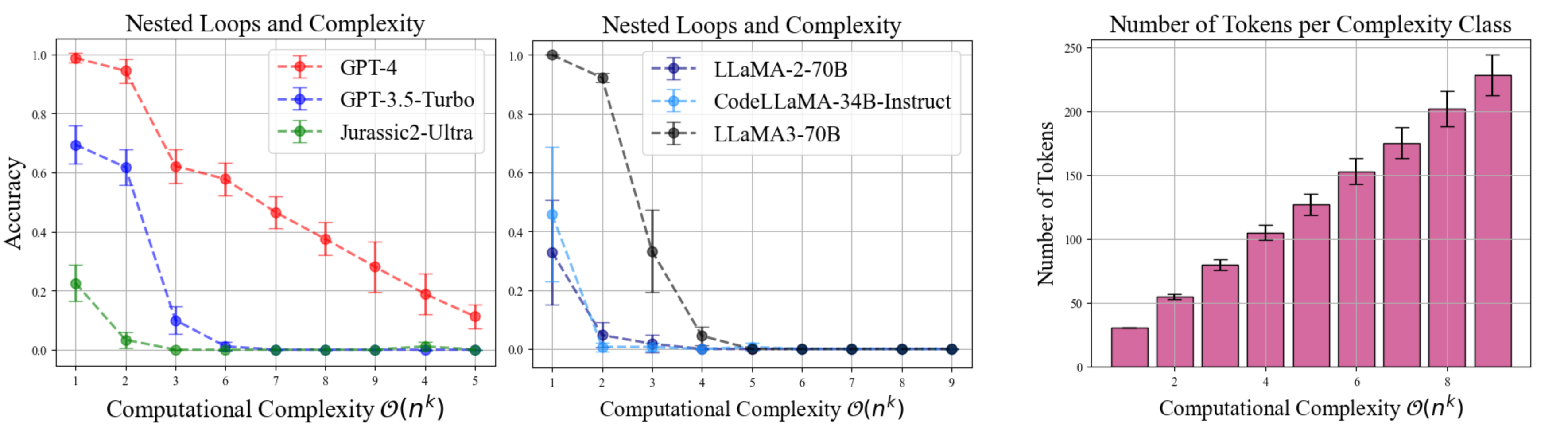}
\caption{Performances of different LLMs on nested loops with increasing computational complexity. On the right, the number of input tokens per complexity class grows linearly.}
\label{fig:complexity-error-res}
\vspace{-4mm}
\end{figure*}
\textbf{Nested loops} are a common instance of programs with polynomial running time $\mathcal{O}(n^k)$, where $k$ is the depth of loop nesting: see, e.g., Figure~\ref{fig:nested-loops}.
In this section, we prompt an LLM with programs that consist of $k$ nested loops with $n$ instructions each, i.e., with time complexity that ranges from $\mathcal{O}(n^{k=1})$ (linear) to $\mathcal{O}(n^{k=9})$. We measure the performance of a model to predict the exact result of the computation. 
By construction, the return value is an integer bounded between $\pm 2^k$, with overall upper- and lower-bounds bounded between $\pm 1024$, so we prevent high-order magnitude operands from influencing an LLM's performances. 
We run $3$ independent runs with batches of $30$ programs each and report the results for GPT-4, GPT-3.5-Turbo, Jurassic2-Ultra, Llama-3-70B, Llama-2-70B and CodeLLaMA-34b-Instruct.

Results in Figure~\ref{fig:complexity-error-res} evidence a \textbf{strong non-linear negative correlation} between the accuracy of GPT-3.5-Turbo, GPT-4 and Llama-3-70B and the computational complexity of the function (right). In contrast, a strong linear correlation characterises the complexity of a function and its length (left). 
For high-performing LLMs (e.g., GPT-3.5, GPT-4, and Llama-3-70B), algorithms whose complexity is beyond quadratic induce the most significant drop in performance. This suggests that the current state-of-the-art models cannot reliably simulate routines whose complexity is cubic or beyond. 
This phenomenon necessitates further investigations to connect the work in~\cite{zhou2023algorithms}, or other works on the computational capabilities of Transformers~\cite{weiss2021thinking}, with the computational complexity of a routine.  
Interestingly, Llama-3-70B was the best-performing model on the straight-line, approximate and critical path code, yet on nested loops, GPT-4 outperforms it by a solid margin.
By inspecting the log results, we noticed that GPT-4, for high complexity programs (i.e., beyond $\mathcal{O}(n^2)$) implicitly unrolls the loops and \textbf{correctly guesses} the final result via \textbf{pattern matching}, surpassing any other model performance, including Llama-3-70B. 

\begin{figure}
\centering
\includegraphics[width=1\linewidth]{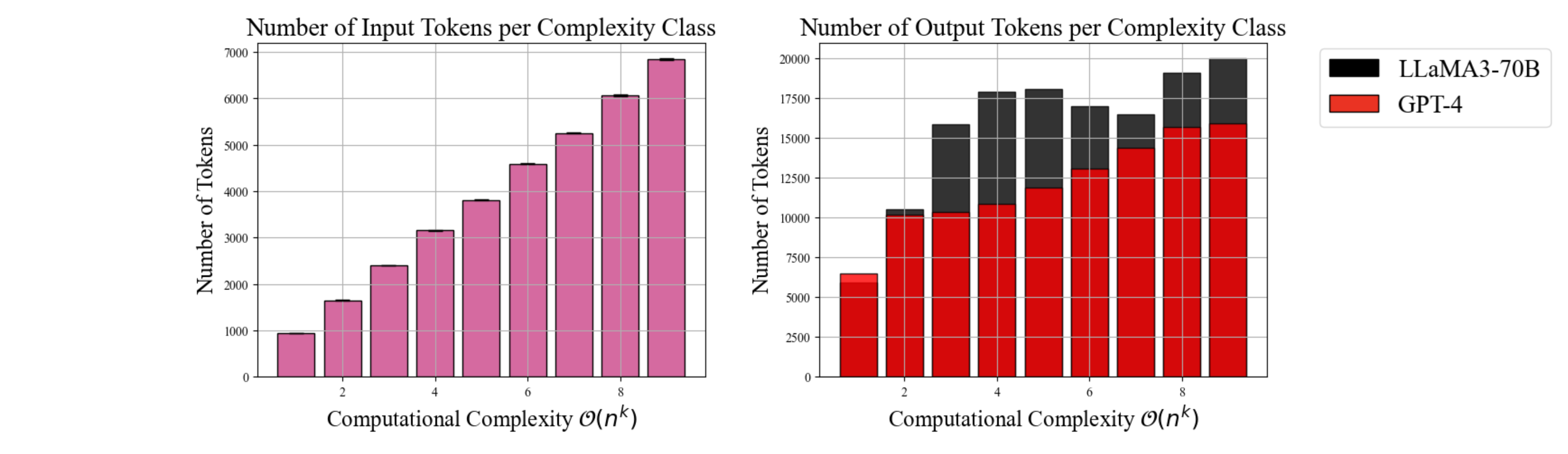}
\caption{On sorting algorithms, the number of input tokens grows linearly (left). At the same time, GPT-4 outputs fewer tokens than Llama-3-70B, especially for complexity larger than $\mathcal{O}(n^2)$. The number is approximately the same for linear complexity. Both the graphs report the \textbf{cumulative} number of tokens over $30$ experiments.}
\label{fig:nested-loops-tokens}
\end{figure}

To give empirical evidence that GPT-4 does implicit computation without unrolling the loops, while Llama-3-70B tries to execute each instruction sequentially, we computed the number of tokens each model outputs in response to programs with different computational complexity.
As reported in Figure~\ref{fig:nested-loops-tokens}, the \textbf{cumulative} number of input tokens grows linearly (left). At the same time, GPT-4 outputs fewer tokens than Llama-3-70B, especially for complexity larger than $\mathcal{O}(n^2)$. The number is approximately the same for linear and quadratic complexity. 

\subsection{Sorting}\label{a:sorting}
\begin{figure*}
    \centering
    \includegraphics[width=1\linewidth]{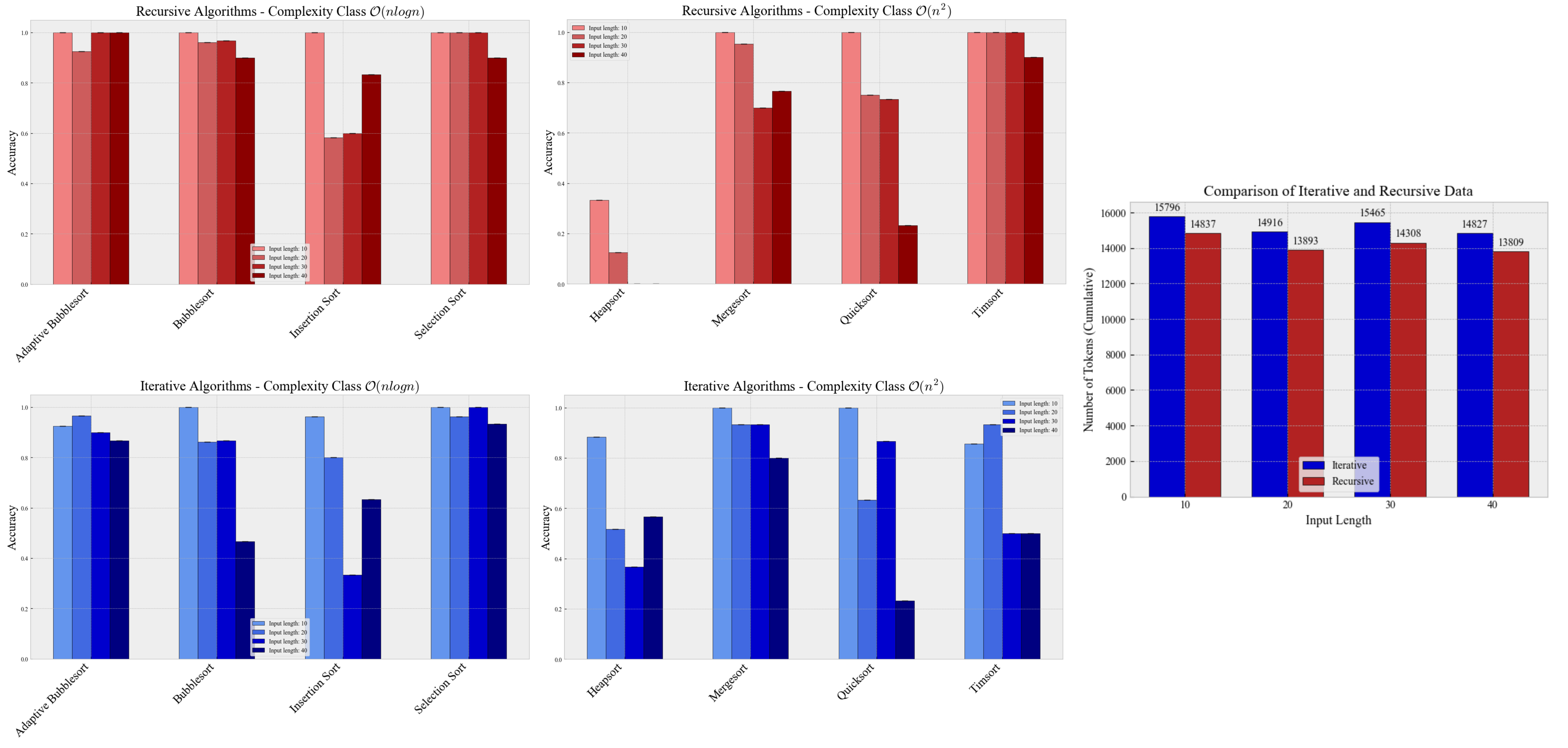}
    \caption{On top, results of GPT-4-Turbo with Chain of Thought prompting technique on different sorting algorithms, both in their recursive (top) and iterative (bottom) versions. Differently from GPT-3.5-Turbo, GPT-4 forces a model to simulate a routine and does not suffer from ``lazy execution'' for longer input vectors (as illustrated in Figure~\ref{fig:sort-iterative} and the relative section).}
    \label{fig:sort-iterative-gpt-cot}
\end{figure*}

\begin{figure*}
    \centering
    \includegraphics[width=1\linewidth]{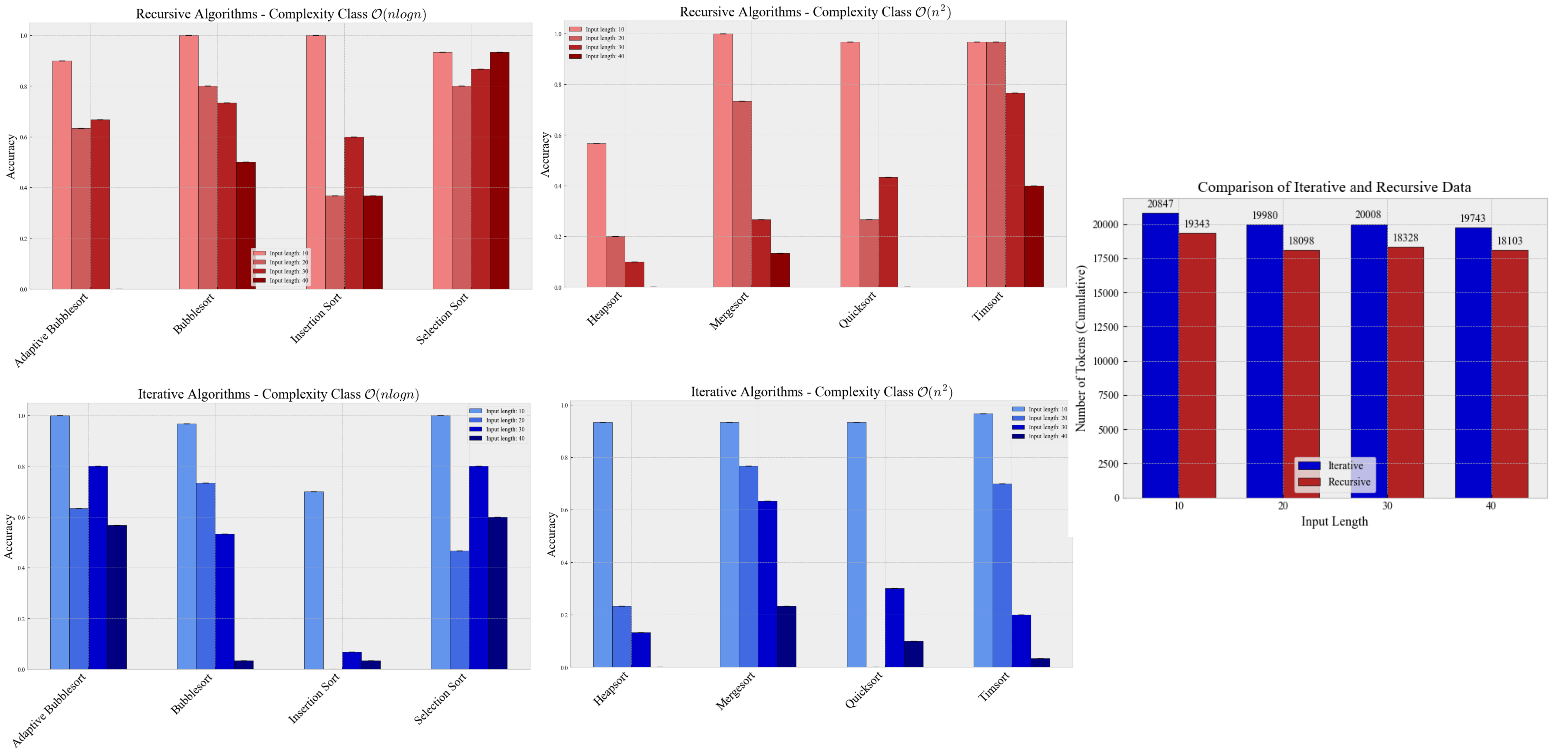}
    \caption{Results of Llama-3-70B with CoT prompting technique on different sorting algorithms, both in their recursive (top) and iterative (bottom) versions. Llama-3-70B is not a good simulator for sorting algorithms and, in general, does not understand the underlying task is sorting.}
    \label{fig:sort-iterative-llama-cot}
\end{figure*}

We report details on each sorting algorithm's space and time complexity in Table~\ref{tab:sorting-appendix}, while results for GPT-4 and Llama-3-70B on all the sorting routines are reported in Figure~\ref{fig:sort-iterative-gpt-cot} and~\ref{fig:sort-iterative-llama-cot}.
\begin{table*}
\centering
\caption{Sorting algorithms space and time complexity. They reference results in Figure~\ref{fig:sort-iterative}.}
\begin{adjustbox}{width=\textwidth} 
\begin{tabular}{|l|l|l|l|l|}
\hline
\textbf{Algorithm} & \textbf{Worst Time Complexity} & \textbf{Average Time Complexity} & \textbf{Best Time Complexity} & \textbf{Space Complexity} \\
\hline
\textcolor{black}{Insertion Sort} & $O(n^2)$ & $\Theta(n^2)$ & $\Omega(n)$ & It: $O(1)$ Rec: $O(n)$ \\
\hline
\textcolor{black}{Selection Sort}  & $O(n^2)$ & $\Theta(n^2)$ & $\Omega(n^2)$ & It: $O(1)$ Rec: $O(n)$ \\
\hline
\textcolor{black}{Bubblesort}  & $O(n^2)$ & $\Theta(n^2)$ & $\Omega(n^2)$ & It: $O(1)$ Rec: $O(n)$ \\
\hline
\textcolor{black}{Adaptive Bubblesort}  & $O(n^2)$ & $\Theta(n^2)$ & $\Omega(n)$ & It: $O(1)$ Rec: $O(n)$ \\
\hline
\textcolor{black}{Quicksort} & $O(n^2)$ & $\Theta(n \log(n))$ & $\Omega(n \log(n))$ & It: $O(n)$ Rec: $O(n)$ \\
\hline
\textcolor{black}{Mergesort}  & $O(n \log(n))$ & $\Theta(n \log(n))$ & $\Omega(n \log(n))$ & It: $O(n)$ Rec: $O(n)$ \\
\hline
\textcolor{black}{Timsort}  & $O(n \log(n))$ & $\Theta(n \log(n))$ & $\Omega(n)$ & It: $O(1)$ Rec: $O(n)$ \\
\hline
\textcolor{black}{Heapsort}  & $O(n \log(n))$ & $\Theta(n \log(n))$ & $\Omega(n \log(n))$ & It: $O(1)$ Rec: $O(\log n)$ \\
\hline
\end{tabular}\label{tab:sorting-appendix}
\end{adjustbox}
\end{table*}


\paragraph{Repetita non iuvant.}
We report a case of emblematic failure that appears frequently with LLMs such as GPT-3.5-Turbo and GPT-4. 
\begin{lstlisting}[escapechar=\%]
[85, %\textbf{\textcolor{magenta}{58}}%, 6, %\textbf{\textcolor{magenta}{58}}%, 34, %\textbf{\textcolor{magenta}{58}}%, 93, 47, 5, 89, 86, 12, 51, 76, 0, 3, 63, 6, 74, 52, 46, 61, 34, 92, 50, 56, 21, 25, %\textbf{\textcolor{magenta}{58}}%, 80].
\end{lstlisting}
The previous sequence, alongside the code for Bubble Sort, is fed to GPT-4. The input vector contains the number $58$ four times, highlighted in \textbf{\textcolor{magenta}{magenta}}). While GPT-3.5-Turbo correctly sorts the input vector, it reports the value $58$ thrice. The LLM output is reported below, with the sequence of interest highlighted in \textbf{\textcolor{magenta}{magenta}}:
\begin{lstlisting}[escapechar=\%]
[0, 3, 5, 6, 6, 12, 21, 25, 34, 34, 46, 47, 50, 51, 52, 56, %\textbf{\textcolor{magenta}{58}}%, %\textbf{\textcolor{magenta}{58}}%, %\textbf{\textcolor{magenta}{58}}%, 61, 63, 74, 76, 80, 85, 86, 89, 92, 93]
\end{lstlisting}
Our hypothesis is that the probability of a sequence that contains the number $58$ four times is so low, conditioned on what has been generated so far, that the model skips one of them, even though we set the presence penalty value to zero.
In this case, which is not isolated and appears frequently for similar inputs,  code simulation and correctness are in tension with the LLM output's probability distribution. 
We hypothesise that the probability of a sequence that contains repeated elements is so low, conditioned on what has been generated so far, that a model skips repeating elements, even when we set the `presence penalty' value to zero.\footnote{\url{https://platform.openai.com/docs/guides/text-generation/frequency-and-presence-penalties}}

\clearpage

\section{Eliciting Code Simulation}\label{a:cosm}
Below, we report the exact implementation used for Chain of Simulation (CoSm).
\begin{lstlisting}
"""
@code@
# 1. Simulate the above program instruction by instruction.
# 2. Report the trace of the program at the end of each iteration.
# 3. Think step by step and reply with the output of the function for the following input: @input@.
"""
\end{lstlisting}

\clearpage

\section{Algorithms Implementation}
\subsection{Sorting Algorithms}\label{a:sorting-code}
\subsubsection{Recursive Algorithms}
\noindent  Insertion Sort:
\begin{lstlisting}
def main(array, size, start=0):
    if start >= len(array) - 1:
        return array
    min_index = start
    for j in range(start + 1, len(array)):
        if array[j] < array[min_index]:
            min_index = j
    array[start], array[min_index] = array[min_index], array[start]
    return main(array, size, start + 1)
\end{lstlisting}

\noindent  Bubble Sort:
\begin{lstlisting}
def main(list_data, length) :
    for i in range(length - 1):
        if list_data[i] > list_data[i + 1]:
            list_data[i], list_data[i + 1] = list_data[i + 1], list_data[i]
    return list_data if length<2 else main(list_data, length - 1)
\end{lstlisting}

\noindent  Selection Sort:
\begin{lstlisting}
def main(array, size, start=0):
    if start >= len(array) - 1:
        return array
    min_index = start
    for j in range(start + 1, len(array)):
        if array[j] < array[min_index]:
            min_index = j
    array[start], array[min_index] = array[min_index], array[start]
    return main(array, size, start + 1)
\end{lstlisting}

\noindent  Adaptive Bubblesort:
\begin{lstlisting}
def main(list_data, length) :
    swapped = False
    for i in range(length - 1):
        if list_data[i] > list_data[i + 1]:
            list_data[i], list_data[i + 1] = list_data[i + 1], list_data[i]
            swapped = True
    return list_data if not swapped else main(list_data, length - 1)
\end{lstlisting}

\noindent  Quicksort:
\begin{lstlisting}
def main(array, high, low=0):
    if high==len(array):
        high=high-1
    if low < high:
        pi = f1(array, low, high)
        main(array,  pi - 1, low)
        main(array, high, pi + 1)
    return array
 
def f1(array, low, high):
    pivot = array[high]
    i = low - 1
    for j in range(low, high):
        if array[j] <= pivot:
            i = i + 1
            (array[i], array[j]) = (array[j], array[i])
    (array[i + 1], array[high]) = (array[high], array[i + 1])
    return i + 1
\end{lstlisting}

\noindent  Merge Sort:
\begin{lstlisting}
def main(arr, r, l=0):
    if r==len(arr):
        r=r-1
    if l < r:
        m = l+(r-l)//2
        main(arr, m, l)
        main(arr, r, m+1)
        f1(arr, l, m, r)
    return arr

def f1(arr, l, m, r):
    n1 = m - l + 1
    n2 = r - m
    L = [0] * (n1)
    R = [0] * (n2)
    for i in range(0, n1):
        L[i] = arr[l + i]
    for j in range(0, n2):
        R[j] = arr[m + 1 + j]
    i = 0
    j = 0
    k = l
    while i < n1 and j < n2:
        if L[i] <= R[j]:
            arr[k] = L[i]
            i += 1
        else:
            arr[k] = R[j]
            j += 1
        k += 1
    while i < n1:
        arr[k] = L[i]
        i += 1
        k += 1
    while j < n2:
        arr[k] = R[j]
        j += 1
        k += 1
\end{lstlisting}

\noindent  Tim Sort:
\begin{lstlisting}
def main(lst, size):
    length = len(lst)
    runs, s_runs = [], []
    new_run = [lst[0]]
    s_array = []
    i = 1
    while i < length:
        if lst[i] < lst[i - 1]:
            runs.append(new_run)
            new_run = [lst[i]]
        else:
            new_run.append(lst[i])
        i += 1
    runs.append(new_run)
    for run in runs:
        s_runs.append(f2(run))
    for run in s_runs:
        s_array = f1(s_array, run)
    return s_array

def f1(left, right):
    if not left:
        return right
    if not right:
        return left
    if left[0] < right[0]:
        return [left[0], *f1(left[1:], right)]
    return [right[0], *f1(left, right[1:])]

def f2(lst):
    length = len(lst)
    for index in range(1, length):
        value = lst[index]
        pos = f3(lst, value, 0, index - 1)
        lst = lst[:pos] + [value] + lst[pos:index] + lst[index + 1 :]
    return lst

def f3(lst, item, start, end):
    if start == end:
        return start if lst[start] > item else start + 1
    if start > end:
        return start
    mid = (start + end) // 2
    if lst[mid] < item:
        return f3(lst, item, mid + 1, end)
    elif lst[mid] > item:
        return f3(lst, item, start, mid - 1)
    else:
        return mid
\end{lstlisting}

\noindent  Heap Sort:
\begin{lstlisting}
def main(u_arr,size):
    n = len(u_arr)
    for i in range(n // 2 - 1, -1, -1):
        f1(u_arr, i, n)
    for i in range(n - 1, 0, -1):
        u_arr[0], u_arr[i] = u_arr[i], u_arr[0]
        f1(u_arr, 0, i)
    return u_arr

def f1(u_arr, index, heap_size):
    largest = index
    left_index = 2 * index + 1
    right_index = 2 * index + 2
    if left_index < heap_size and u_arr[left_index] > u_arr[largest]:
        largest = left_index

    if right_index < heap_size and u_arr[right_index] > u_arr[largest]:
        largest = right_index

    if largest != index:
        u_arr[largest], u_arr[index] = u_arr[index], u_arr[largest]
        f1(u_arr, largest, heap_size)
\end{lstlisting}

\subsubsection{Iterative Algorithms}
\noindent  Insertion Sort:
\begin{lstlisting}
def main(arr, size):
    for j, val in enumerate(arr[1:]):
        i = j
        while j >= 0 and val < arr[j]:
            arr[j + 1] = arr[j]
            j -= 1
        if j != i:
            arr[j + 1] = val
    return arr
\end{lstlisting}

\noindent  Bubble Sort:
\begin{lstlisting}
def main(collection, size=0):
    length = len(collection)
    for i in reversed(range(length)):
        for j in range(i):
            if collection[j] > collection[j + 1]:
                collection[j], collection[j + 1] = collection[j + 1], collection[j]
    return collection
\end{lstlisting}

\noindent  Selection Sort:
\begin{lstlisting}
def main(collection, size=0):
    length = len(collection)
    for i in reversed(range(length)):
        for j in range(i):
            if collection[j] > collection[j + 1]:
                collection[j], collection[j + 1] = collection[j + 1], collection[j]
    return collection
\end{lstlisting}

\noindent  Adaptive Bubblesort:
\begin{lstlisting}
def main(collection, size=0):
    length = len(collection)
    for i in reversed(range(length)):
        swapped = False
        for j in range(i):
            if collection[j] > collection[j + 1]:
                swapped = True
                collection[j], collection[j + 1] = collection[j + 1], collection[j]
        if not swapped:
            break
    return collection
\end{lstlisting}

\noindent  Quicksort:
\begin{lstlisting}
def main(arr, h, l=0):
    if h==len(arr):
        h=h-1
    size = h - l + 1
    stack = [0] * (size)
    top = -1
    top = top + 1
    stack[top] = l
    top = top + 1
    stack[top] = h
    while top >= 0:
        h = stack[top]
        top = top - 1
        l = stack[top]
        top = top - 1
        p = f1( arr, l, h )
        if p-1 > l:
            top = top + 1
            stack[top] = l
            top = top + 1
            stack[top] = p - 1
        if p + 1 < h:
            top = top + 1
            stack[top] = p + 1
            top = top + 1
            stack[top] = h
    return arr
\end{lstlisting}

\noindent  Merge Sort:
\begin{lstlisting}
def main(a, size):
    width = 1   
    n = len(a)                                         
    while (width < n):
        l=0;
        while (l < n):
            r = min(l+(width*2-1), n-1)        
            m = min(l+width-1,n-1)           
            f1(a, l, m, r)
            l += width*2
        width *= 2
    return a

def f1(a, l, m, r):
    n1 = m - l + 1
    n2 = r - m
    L = [0] * n1
    R = [0] * n2
    for i in range(0, n1):
        L[i] = a[l + i]
    for i in range(0, n2):
        R[i] = a[m + i + 1]
    i, j, k = 0, 0, l
    while i < n1 and j < n2:
        if L[i] <= R[j]:
            a[k] = L[i]
            i += 1
        else:
            a[k] = R[j]
            j += 1
        k += 1
    while i < n1:
        a[k] = L[i]
        i += 1
        k += 1
    while j < n2:
        a[k] = R[j]
        j += 1
        k += 1
\end{lstlisting}

\noindent  Tim Sort:
\begin{lstlisting}
def main(arr,n):
    min_run = 32
    n = len(arr)
    for i in range(0, n, min_run):
        f2(arr, i, min((i + min_run - 1), n - 1))
    size = min_run
    while size < n:
        for start in range(0, n, size * 2):
            middle = min((start + size - 1), (n - 1))
            end = min((start + size * 2 - 1), (n - 1))
            if middle < end:
                f1(arr, start, middle, end)
        size *= 2
    return arr
    
def f2(arr, left=0, right=None):
    if right is None:
        right = len(arr) - 1
    for i in range(left + 1, right + 1):
        key_item = arr[i]
        j = i - 1
        while j >= left and arr[j] > key_item:
            arr[j + 1] = arr[j]
            j -= 1
        arr[j + 1] = key_item

def f1(arr, left, middle, right):
    if arr[middle] <= arr[middle + 1]:
        return
    left_copy = arr[left:middle + 1]
    right_copy = arr[middle + 1:right + 1]
    left_copy_index = 0
    right_copy_index = 0
    s_index = left
    while left_copy_index < len(left_copy) and right_copy_index < len(right_copy):
        if left_copy[left_copy_index] <= right_copy[right_copy_index]:
            arr[s_index] = left_copy[left_copy_index]
            left_copy_index += 1
        else:
            arr[s_index] = right_copy[right_copy_index]
            right_copy_index += 1
        s_index += 1
    while left_copy_index < len(left_copy):
        arr[s_index] = left_copy[left_copy_index]
        left_copy_index += 1
        s_index += 1
    while right_copy_index < len(right_copy):
        arr[s_index] = right_copy[right_copy_index]
        right_copy_index += 1
        s_index += 1
\end{lstlisting}

\noindent  Heap Sort:
\begin{lstlisting}
def main(arr, n):
    f1(arr, n)
    for i in range(n - 1, 0, -1):
        arr[0], arr[i] = arr[i], arr[0]
        j, index = 0, 0
        while True:
            index = 2 * j + 1
            if (index < (i - 1) and
                arr[index] < arr[index + 1]):
                index += 1
            if index < i and arr[j] < arr[index]:
                arr[j], arr[index] = arr[index], arr[j]
            j = index
            if index >= i:
                break
    return arr

def f1(arr, n):
    for i in range(n):
        if arr[i] > arr[int((i - 1) / 2)]:
            j = i
            while arr[j] > arr[int((j - 1) / 2)]:
                (arr[j],
                 arr[int((j - 1) / 2)]) = (arr[int((j - 1) / 2)],arr[j])
                j = int((j - 1) / 2)
\end{lstlisting}

\subsection{Standard Algorithms and Variations}\label{a:classic-algs}
\noindent Fibonacci (iterative):
\begin{lstlisting}
def f(n):
    a, b = 0, 1
    if n <=1:
        return n       
    else:
        for i in range(1, n):
            c = a + b
            a = b
            b = c
        return b
\end{lstlisting}

\noindent Padovan (iterative):
\begin{lstlisting}
def g(n):
    a, b = 1, 1
    c, d = 1, 1
    for i in range(3, n+1):
        d = a + b
        a = b
        b = c
        c = d 
    return d
\end{lstlisting}

\noindent Bubble Sort (iterative):
\begin{lstlisting}
def f(v):
    n = len(v)
    for i in range(n):
        for j in range(0, n-i-1):
            if v[j] > v[j+1]:
                v[j], v[j+1] = v[j+1], v[j]
    return v
\end{lstlisting}

\noindent Bubble Sort Descending (iterative):
\begin{lstlisting}
def g(v):
    n = len(v)
    for i in range(n):
        for j in range(0, n-i-1):
            if 0 > v[j] - v[j+1]:
                v[j], v[j+1] = v[j+1], v[j]
    return v
\end{lstlisting}

\noindent Gauss Sum:
\begin{lstlisting}
def f(n):
    tot = 0
    for i in range(n):
        tot += i
    return tot
\end{lstlisting}

\noindent Gauss Sum and Subtraction:
\begin{lstlisting}
def g(n):
    tot = 0
    for i in range(n):
        tot += (i if i%2==0 else -i)
    return tot
\end{lstlisting}

\noindent Is Prime:
\begin{lstlisting}
def f(n):
    if n < 2: return False
    for x in range(2, int(n**0.5) + 1):
        if n % x == 0:
            return False
    return True
\end{lstlisting}

\noindent Is Prime on Successor:
\begin{lstlisting}
def g(n):
    n = n+1
    if n < 2: return False
    for x in range(2, int(n**0.5) + 1):
        if n % x == 0:
            return False
    return True
\end{lstlisting}

\noindent Collatz Sum Even:
\begin{lstlisting}
    def g(n):
        s = n
        while n != 1:
            if n % 2 == 0:
                n = n // 2
                s += n
            else:
                n = 3 * n + 1
        return s
\end{lstlisting}

\end{document}